\pdfoutput=1
\documentclass[10pt,twocolumn,letterpaper]{article}
\usepackage[pagenumbers]{cvpr}

\usepackage[T1]{fontenc}
\usepackage{graphicx}
\usepackage{amsmath}
\usepackage{amssymb}
\usepackage{booktabs}
\usepackage{multirow}
\usepackage{array}
\usepackage[dvipsnames]{xcolor}
\usepackage{colortbl}
\definecolor{oursbg}{gray}{0.92}
\newcommand{\ours}{\rowcolor{oursbg}}

\usepackage{tikz}
\usetikzlibrary{arrows.meta,positioning,fit,calc,shapes.geometric,shapes.misc,backgrounds}
\usepackage{xspace}

\definecolor{exactgreen}{rgb}{0.10,0.55,0.25}
\definecolor{failorange}{rgb}{0.85,0.40,0.05}

\newcommand{\pre}[2][]{\ifx\relax#2\relax{\color{NavyBlue}[#1]}\else{\color{NavyBlue}#2}\fi}

\newcommand{\sysname}{Abra-CAD\xspace}

\DeclareMathOperator{\IoU}{IoU}

\graphicspath{{figures/}}

\extrafloats{100}

\definecolor{cvprblue}{rgb}{0.21,0.49,0.74}
\usepackage[pagebackref,breaklinks,colorlinks,allcolors=cvprblue]{hyperref}

\def\confName{CVPR}
\def\confYear{2027}

\title{ABRA-CAD: Geometry-Guided Reconstruction of CAD Solids and Construction Programs}

\author{Anush Bharathi \qquad Ramanathan Muthuganapathy\\
Indian Institute of Technology Madras}

\makeatletter
\apptocmd{\@maketitle}{\par\noindent\begin{minipage}{\textwidth}
\centering
\includegraphics[width=\textwidth]{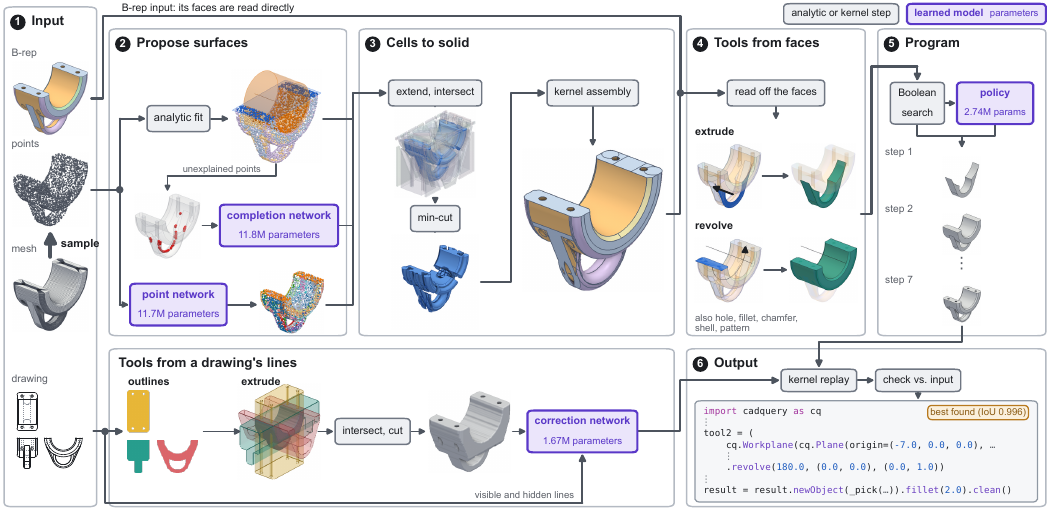}
\captionof{figure}{\textbf{From an observation to a CadQuery program.}
From a mesh or points (1), an analytic fit and a point network
propose surfaces (2; the completion network shown is used only in an
ablation), and a min-cut over the cells
they bound gives a solid (3). Construction tools are read off that
solid's faces, or directly off a B-rep (4), and Boolean search
followed by the learned policy finds a program (5). A drawing is
carved from its views and refined by an auxiliary correction network. Every route
ends in a program that is replayed in the kernel, checked against the
input and written out as stand-alone CadQuery (6; the program shown
reaches IoU 0.996). Grey boxes are analytic or kernel steps; tinted
boxes are learned models with their parameter counts.}
\label{fig:pipeline}
\end{minipage}\par\vspace{4pt}
}{}{}
\makeatother

\begin{document}
\maketitle
\begin{abstract}
Engineering parts are often shared as meshes, drawings or CAD files
without their construction history, so they cannot be edited the way
they were designed. Recovering an editable CAD program means finding
both the modeling operations and their dimensions. Recent methods train
large code generators to predict both. We observe that a boundary
representation (B-rep) already contains most of these dimensions: a
face outline can serve as an extrusion profile, and a cylindrical face
as a hole. \sysname therefore reads candidate tools directly off the
faces and searches only for which tools to combine, and in what order.
A fast Boolean search tests combinations on points inside and outside
the part before running the CAD kernel, and a small learned policy
finds programs that this search misses. Meshes and point clouds are
first rebuilt into a B-rep by fitting surfaces such as planes and
cylinders. When search cannot match the input, the method returns a
program that extrudes cross-sections of the mesh. Drawings are carved
into a program from their three views. On 748 ABC parts, our programs
reach 0.864 mean normalized volume IoU from meshes with
this fallback (0.648 from search alone), against 0.672 for
CADENA, and 0.744 from drawings, against 0.334 for Ortho2CAD RL. Given
the B-rep itself, our programs rebuild 403 of the 748 parts exactly. Surface- and
feature-level measures reveal details that volume overlap misses.
\end{abstract}

\section{Introduction}
\label{sec:intro}

A STEP file, the standard exchange format for boundary representations
(B-reps), can preserve the geometry of a mechanical part precisely
while omitting the operations that created it. A mesh or scan likewise
records the final shape, but not the sketches, cuts, and features that
an engineer would modify to revise the design. Recovering an editable
CAD program is therefore a different problem from reconstructing a
solid. It requires finding operations that recreate the geometry,
without assuming that the original construction history is unique.
For designs already represented in an engineering library, retrieving a
suitable STEP model and recovering its construction offers another
route to editable CAD without generating the geometry from scratch.

Recent approaches generate CAD programs from point clouds, meshes,
and images~\cite{rukhovich2025cadrecode,kolodiazhnyi2026cadrille,
kabisov2026cadena,nehme2026cadfit,nehme2026stepcad}.
These methods have expanded the range of recoverable constructions,
but they still face a difficult joint inference problem: determining
\emph{which} operations created the part and \emph{what geometry}
each operation should use. Code-generating models predict operations together with their numbers;
geometry-guided methods estimate candidate sketches and dimensions, then
test them. Estimating dimensions is unavoidable when only sampled
geometry is available. A B-rep offers more. The outline of a
planar face can define an extrusion profile; another face can fix its
extent; a cylindrical face can determine a hole cutter. \emph{The
construction sequence is missing, but the geometry of its candidate
tools is already present.}

We introduce \sysname (Analytic B-Rep Assisted CAD Generator), which extends this
observation, used for face extrusions and revolves in prior
work~\cite{willis2021fusion360,xu2021zonegraph,yin2025rlcad}, to holes,
fillets, chamfers, shells and defeatured bodies. Rather than predicting numerical parameters for
supported operations from an observed B-rep, we construct tools
directly from its faces and search for how to combine them. The
remaining decisions are discrete: which tools to use, which Boolean
operations to apply, and in what order. Boolean search first tests many candidate
combinations on points inside and outside the target, and runs the CAD
kernel only on promising programs. A small, 2.74M-parameter learned policy searches complementary
construction sequences in order of their conditional code
length, the bits the policy needs to encode their choices given the target. 
The recovered operations are exported as executable CadQuery programs 
with explicit geometric parameters. On the Fusion~360 Gallery test split,
\sysname recovers volume-exact constructions for 86.5\% of 1,725
designs within 100 kernel evaluations, compared with 67.5\% reported
for the published Gallery agent in its own
environment~\cite{willis2021fusion360}.

The same principle extends to less complete inputs. From a mesh or point
cloud, \sysname first rebuilds a B-rep and then searches it the same way;
when that search fails, it returns a program that extrudes cross-sections
of the mesh. A drawing is carved into a program from its three views.
These inputs carry less geometric certainty than a B-rep. Accordingly, we evaluate not only
whether a program executes or its solid overlaps the target, but also
whether the reconstructed boundary retains the source's geometric
features.

\noindent\textbf{Our contributions are:}
\begin{itemize}
    \item \textbf{Probe-screened search over an extended target-derived tool set:}
    Building on face-derived extrusions~\cite{willis2021fusion360,xu2021zonegraph},
    we derive holes, fillets, chamfers, shells and defeatured bodies from
    B-rep faces, and screen Boolean combinations of these tools on
    inside/outside probe points before any kernel replay. Volume-exact
    programs are emitted as stand-alone CadQuery scripts.
    \item \textbf{A complementary learned policy:}
    A compact policy, trained on its own successful constructions,
    recovers programs that Boolean search misses.
    \item \textbf{The same reading and search for other inputs:}
    From meshes and point clouds we rebuild a B-rep and search its
    tools in the same way, and from orthographic drawings we carve a
    program from the three views. A section-stack fallback and a
    learned drawing correction are auxiliary steps; we report results
    with and without them for meshes and drawings.
    \item \textbf{Evaluating geometric and feature fidelity:}
    We evaluate reconstructed CAD programs beyond volume overlap,
    measuring how well they preserve the target's surfaces and
    geometric features, including small details that conventional
    volume-based metrics can overlook.
\end{itemize}

\section{Related work}
\label{sec:related}

\paragraph{Reconstruction from incomplete geometry.}
Recovering a CAD model from samples can mean recovering its boundary
or recovering a sequence of operations. PolyFit, Point2CAD, and
Split-and-Fit infer analytic surfaces and B-rep structure from geometric
observations~\cite{nan2017polyfit,liu2024point2cad,liu2024splitfit},
but do not reconstruct the sequence of modeling operations. CAD-SIGNet,
CAD-Diffuser, SfmCAD and CADCrafter instead infer construction sequences
from points, voxels and
images~\cite{khan2024cadsignet,ma2024cadiffuser,li2024sfmcad,chen2025cadcrafter}; CAD-Recode and
cadrille generate CAD commands from sampled geometry, and CADFS
generates FeatureScript programs from text and
images~\cite{rukhovich2025cadrecode,kolodiazhnyi2026cadrille,pyatov2026cadfs}.
More recent methods use geometric feedback to improve predictions:
CADENA and StepCAD execute and refine candidate programs, while
CADFit fits sketch-based operations directly to a
mesh~\cite{kabisov2026cadena,nehme2026stepcad,nehme2026cadfit}.
These methods recover editable constructions from incomplete inputs,
where both the operation sequence and its numerical geometry must be
inferred.

\paragraph{Recovering programs from observed boundaries.}
When the boundary is known, it can constrain the construction
itself. InverseCSG searches Boolean combinations of primitives;
B-rep-to-CSG conversion and zone graphs use existing boundaries to
constrain the possible decompositions~\cite{du2018inversecsg,shapiro1991csg,xu2021zonegraph}.
The Fusion~360 Gallery agent and RLCAD learn construction actions
defined by target faces~\cite{willis2021fusion360,yin2025rlcad}, while BrepCoder learns to reason
about B-reps and code, and BrepGiff and CADSketcher generate B-reps and
parametric sketches~\cite{kim2026brepcoder,guo2025brepgiff,liu2026cadsketcher}. CADParser, Brep2Seq and CADCL map B-reps directly to construction
sequences~\cite{zhou2023cadparser,zhang2024brep2seq,liang2025cadcl}.
Pointer-CAD brings face and edge references into sequential CAD generation, grounding
new operations in the intermediate B-rep~\cite{qi2026pointercad}.
\sysname shares this use of the target boundary. It derives a broader
set of tools from the target's faces and screens their Boolean
combinations on probe points before executing any program in the
kernel. Execution-guided
program synthesis and expert iteration inform the learned search
component~\cite{ellis2019repl,ellis2021dreamcoder,anthony2017exit};
the recovered program is then executed and its geometry compared
against the target, rather than judged executable alone.

\paragraph{Reconstruction from orthographic drawings.}
Orthographic views constrain outlines and internal features, but
leave depth and cross-view correspondences ambiguous. PlankAssembly
and CAD2Program infer constructive models from multiple
views~\cite{hu2023plankassembly,wang2025cad2program}, while Ortho2CAD
produces programs from raster drawing sheets~\cite{joglekar2026ortho2cad}.
Abra-CAD instead combines profile construction from drawing geometry
with the same downstream program execution and evaluation used for
its other inputs.

\paragraph{Retrieval and CAD reuse.}
Retrieval offers a different starting point when a related design
already exists. OrthoCAD-322K retrieves CAD models from orthographic
drawings~\cite{mahajan2025orthocad}; MM-CAD organizes large,
multimodal CAD collections through manifold-aware sampling and
semantically structured subsets~\cite{bharathi2026mmcad}; and CAD-RAG
uses retrieved designs to guide editable
CAD generation~\cite{ananthakrishnan2025cadrag}. These approaches
make existing designs easier to find or reuse. They also motivate a
complementary step: recover a construction program from a retrieved
STEP boundary, rather than regenerate the retrieved shape. Abra-CAD
addresses this boundary-to-program step; integrating it with retrieval
is a future direction.
\section{Methodology}
\label{sec:method}

\sysname connects geometric reconstruction to CAD program recovery
(Fig.~\ref{fig:pipeline}). A STEP boundary representation (B-rep)
already defines a solid through its faces, edges and their connections.
For meshes and point clouds, we first reconstruct a B-rep, then search
for operations that reproduce it. Mesh section stacks also provide
recorded construction programs when search does not fit the input
sufficiently well. Orthographic drawings follow a separate route that
constructs and records operations directly from the projected views.
We evaluate the geometry produced by executing these constructions.

\subsection{Reconstruction objective and action space}
\label{sec:actions}
A construction program $P=(a_1,\ldots,a_k)$ is a sequence of modeling
actions that executes to a solid $S(P)$. For an observed B-rep $T$, we
seek any program satisfying
\begin{equation}
\label{eq:exact_program}
\left|\operatorname{IoU}(S(P),T)-1\right|\leq\varepsilon,
\qquad \varepsilon=10^{-5},
\end{equation}
within a budget of $V$ kernel evaluations, each the replay of a new
program prefix.
We call such a program
\emph{volume-exact}. It need not match the original design history,
since many programs build the same solid. For an incomplete observation
$O$, we first reconstruct a solid $\widehat T(O)$. Program selection
then measures how well the executed solid fits the observed input,
rather than requiring exact agreement with the intermediate reconstruction.

Each B-rep action $a_t=(f,r,o)$ has three parts. The family $f$ is an
extrusion, hole, revolution, fillet, chamfer, shell or pattern. The
references $r$ point to target faces, which fix the tool's profile,
placement and dimensions. For Boolean actions, the effect $o$ is a new body, union,
difference or intersection. Finishing actions (fillet, chamfer, shell) instead modify edges or
faces of the current solid.

\subsection{Reconstruction from incomplete inputs}
\label{sec:entity}
\label{sec:builder}

Unlike a STEP B-rep, a mesh or point cloud does not explicitly identify
the analytic faces of a CAD solid. We fit and propose surfaces, then
reconstruct a solid bounded by them. We also retain section-based
constructions as a fallback.

\paragraph{Analytic surface fitting.}
Most faces of mechanical parts lie on planes, cylinders, cones, spheres
or tori. We fit these primitives to mesh or oriented-point samples
wherever the samples support them. Each fitted surface is kept
as a \emph{carrier}, not trimmed to face boundaries; one plane, for example, can carry
several coplanar faces. Profiles recovered from the samples also give
candidate extrusions and revolutions.

\paragraph{Learned surface proposals.}
Fitting alone misses surfaces when the samples are fragmented or noisy.
An 11.7M-parameter set-prediction network therefore proposes additional
surfaces: PointNet features~\cite{qi2017pointnet} feed a fixed set of
learned DETR-style queries~\cite{carion2020detr}, each predicting at
most one surface with its type, parameters and supporting points. A
second, 11.8M-parameter completion network takes the surfaces already
fitted and the points they explain, and proposes surfaces for the
remaining points. Both use synthetic Zero-to-CAD training
programs~\cite{ataei2026zerotocad}; completion also uses ABC training
parts. The completion network is used only in an ablation; the
reported mesh and point results use only the set-prediction network. Architectures
are in the appendix.

\paragraph{Cell-based B-rep reconstruction.}
Surfaces alone do not say which side is material. We extend each
surface into a box enclosing the part (planes across the whole box,
cylinders and cones somewhat beyond their supporting points), which
splits the box into
closed cells with shared faces. A graph min-cut then labels each cell
as material or empty. Its cost favors boundaries supported by input
points with consistent normals, penalizes the area of unsupported
boundaries, and fixes cells touching the box as empty. Fusing the
material cells in Open CASCADE~\cite{occt} removes their shared internal
faces, and fragments on the same analytic surface are merged wherever
the result stays closed and manifold. After a kernel validity check, the result is a B-rep solid whose
faces lie on the recovered analytic surfaces. It can enter the same
program search as an observed STEP solid (Sec.~\ref{sec:brep}).

\paragraph{Section-stack fallback.}
Some parts resist surface recovery. We therefore also slice the input
mesh at successive levels and approximate each cross-section by closed
profiles of lines, arcs and circles, or by polygons where necessary.
Each profile is extruded to the next level, and the slabs are fused
into a solid. Further candidates combine stacks along different axes or
apply local additions and cuts. Because each extrusion and Boolean operation is recorded, a section
stack can be written as a CadQuery program~\cite{cadquery}. This
auxiliary step provides a geometric-fit fallback, although the resulting sequence of
slabs may be less intuitive to edit than a compact feature-based program.

\paragraph{Candidate and program selection.}
For the mesh results reported here, we rank reconstructed solids by
fit to the input mesh; a fixed-priority selector is studied separately
in the appendix. A separate 2.75M-parameter face policy, similar to
the policy of Sec.~\ref{sec:brep}, adds one candidate program by
searching the selected reconstruction (appendix). We then run B-rep
program search on the chosen solid (Sec.~\ref{sec:brep}). The
recorded construction of a section
stack supplies a fallback, using the best-fitting stack if the chosen
solid has no such record. We replay candidate programs and return the
shortest whose input-mesh fit lies within a fixed tolerance of the best
candidate. The fallback can therefore reconstruct a different solid
from the one initially selected, provided that its input fit is
sufficiently close.

\paragraph{Point clouds.}
On DeepCAD~\cite{wu2021deepcad} and Fusion~360 we follow the 256-point
protocol of prior work and rebuild a mesh from the points. For the ABC program evaluation we instead use 8,192 oriented points,
which are meshed before entering the same reconstruction and program
selection pipeline. These input protocols differ in both density and
normal availability.

\paragraph{Orthographic drawings.}
A drawing gives three 2D views instead of 3D samples, so we carve a
solid from them. We trace outlines and interior lines in each view of the raster sheet,
using printed dimensions, when available, to set the extents. A box at
these extents is intersected with each view's outline, extruded along
its viewing direction; a solid of revolution replaces the extrusions
when the views show one. Closed regions of a view then become profile
or revolved cuts, with depths read from lines in the other views. A cut
is kept only if its re-rendered visible and hidden lines match the
drawing better. Polygon loops whose vertices lie on a circle become
true circles. As an auxiliary step, a 1.67M-parameter network then predicts occupancy
from the drawn lines; where it disagrees with the carve beyond fixed
thresholds, the region is added or removed as a union of boxes. The
recorded operations form the program.

\subsection{Program recovery from observed B-reps}
\label{sec:brep}
\begin{figure*}[!t]
\centering
\includegraphics[width=.84\textwidth]{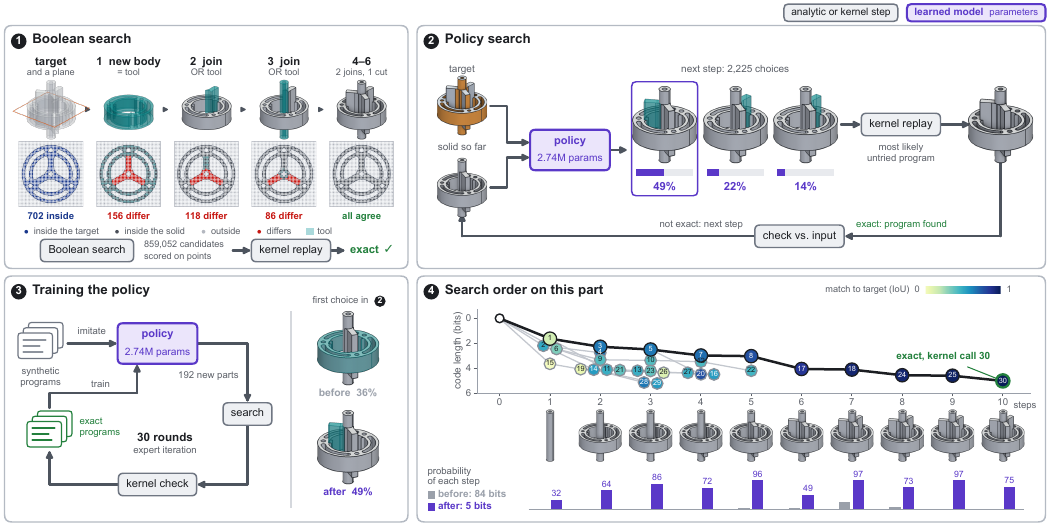}
\caption{\textbf{Two searches using the same target-derived tools.}
On a training part, Boolean search scores 859,052 candidate
combinations on probe points and replays a single six-operation
program in the kernel. The learned policy finds a different
volume-exact program at its 30th kernel evaluation; expert iteration
lowers that program's conditional code length from 84 to 5 bits.}
\label{fig:search_walkthrough}
\end{figure*}

\paragraph{Target-derived tools.}
The target B-rep supplies geometric constraints for a finite set of
candidate construction tools. Planar-face outlines give extrusion profiles and other faces
fix their extents; cylindrical faces give hole cutters; surfaces of
revolution give revolved tools; fillets, chamfers and shells act on the
current solid. For these supported tool families, the target geometry determines
the numerical operands before search. Search chooses which tools to
apply, their Boolean effects and their order. Finishing features can also be removed from the target
temporarily (defeaturing): the simpler body is recovered first, and
later actions restore the features.

\paragraph{Probe-based Boolean search.}
Executing every candidate in the CAD kernel would be slow, so we first
screen candidates on probe points (Fig.~\ref{fig:search_walkthrough},
panel~1). Probes are 1,024 random points in the target's bounding box,
plus point pairs just inside and just outside each target face. Each tool $U$ gets a binary mask
$m(U)$ recording which probes it contains. Unions, intersections and
differences of tools then become bitwise operations on masks. A beam
search keeps the combinations whose masks differ least from the target
mask $m(T)$, weighting probes near faces more heavily; combinations with
identical masks are merged.

Only the best-scoring candidates are replayed in the kernel. Volume
exactness can still miss a small face, so among volume-exact replays the
search prefers one that also places sampled interiors of every target
face on its boundary. A candidate that agrees on every probe is
consistent on probes; only its kernel replay verifies that it rebuilds
the target. When such a
candidate's solid still differs from the target, we add probes in
the missing and excess regions and search again. If this fails, the remaining budget passes to the learned policy.

\paragraph{Policy and expert iteration.}
The complementary search uses a compact 2.74M-parameter graph
transformer over the face graphs of the target and the current solid
(Fig.~\ref{fig:search_walkthrough}, panel~2). At each step it predicts a
construction family and the target faces it references, given what has
already been built. We first train it to imitate synthetic Zero-to-CAD
programs~\cite{ataei2026zerotocad}. Expert
iteration~\cite{anthony2017exit} then lets the policy search fresh
Zero-to-CAD training targets (Fig.~\ref{fig:search_walkthrough},
panel~3); successful volume-exact search trajectories supply new imitation
examples. A Kullback--Leibler (KL) penalty keeps the updated policy close
to the imitation policy. During search, equivalent sibling actions are
pooled and previously reached solids are not revisited.

\paragraph{Program ranking and execution.}
The policy expands partial programs in order of their accumulated
conditional code length (Fig.~\ref{fig:search_walkthrough}, panel~4),
\begin{equation}
\label{eq:policy_code}
b_\pi(P\mid T)=-\sum_{t=1}^{k}\log_2\pi(a_t\mid s_t,T),
\end{equation}
where $s_t$ is the solid before action $a_t$. This is the number of bits
the policy needs to encode the program's choices given the
target~\cite{ellis2021dreamcoder}; a lower value means the program is
more probable under the policy, not that its script is shorter. The reported budget $V$ counts newly constructed program prefixes
replayed in the CAD kernel, not all kernel operations performed by the
system. Tool construction, probe classification, defeaturing and
counterexample generation are excluded from $V$ but contribute to
wall time. Finally, a deterministic emitter expands face-referenced actions
into stand-alone CadQuery scripts with explicit sketch geometry and
dimensions~\cite{cadquery}. We report the success of independent
script replay separately in the appendix.

\subsection{Boundary verification and feature preservation}
\label{sec:boundary}

\paragraph{Oriented surface coverage.}
Volume overlap can stay high while small surface details are lost, so
we also measure surfaces. Adjacent source faces on the same underlying
surface are grouped into patches and sampled by area. A sample is
covered if the output surface has a point within $\tau D_S$, where $D_S$
is the source diameter, with an outward normal within $30^\circ$.
\emph{Oriented Surface Coverage} (OSC) averages the covered fraction
over patches with equal weight, so small patches count as much as large
ones. A separate bidirectional check tests sampled boundary agreement in
both directions; unlike volume exactness, it is a finite-sample
surface test.

\paragraph{Feature preservation.}
The Quiddity feature recognizer~\cite{quiddity} identifies, in the source, the faces that
support recognized features such as holes, fillets and chamfers. A
feature is retained when each of its supporting faces reaches at least
95\% coverage at $\tau=10^{-3}$. FeatR95 is the fraction of recognized source features whose supporting
faces meet this threshold. It tests geometric support, not whether the
output contains the same parametric feature tree. OSC also covers
faces without a recognized feature label.

\section{Experiments}
\label{sec:exp}

We evaluate \sysname in two settings: recovering a construction program
from an observed B-rep, and reconstructing one from incomplete geometric
observations. The experiments compare program fidelity on the same parts,
then examine exact recovery and generalization at scale.

\subsection{Experimental setup}
\label{sec:exp_setup}

\paragraph{Data.}
Our cross-modal evaluation uses 748 single-solid ABC
parts~\cite{koch2019abc}: 400 single-solid files of the 533-file ABC-EVAL set and 348 CADBench
ABC parts (58 from each operation split and face-count tier);
every method is scored on the same parts. For B-rep recovery we additionally evaluate all 1,725 Fusion~360 Gallery
test designs~\cite{willis2021fusion360} and four CADBench
splits~\cite{doris2026cadbench}, with 3,000 designs each.

\paragraph{Inputs and baselines.}
Methods reconstruct the same underlying parts from their respective
inputs. CAD-Recode~\cite{rukhovich2025cadrecode} and
cadrille~\cite{kolodiazhnyi2026cadrille} use 256-point clouds;
our point route uses 8,192 oriented samples. CADENA~\cite{kabisov2026cadena},
StepCAD~\cite{nehme2026stepcad} and CADFit~\cite{nehme2026cadfit}
use meshes, while Ortho2CAD~\cite{joglekar2026ortho2cad} uses
orthographic drawings. We evaluate released baselines on the indicated
inputs and report failures as zero.

\paragraph{Metrics.}
\emph{Exact} denotes a program whose kernel replay matches the
reference volume within $|\operatorname{IoU}-1|\leq10^{-5}$.
\emph{Verified} counts outputs passing the bidirectional boundary check,
independently of volume exactness. We report normalized volume IoU
(IND, volume IoU after normalizing each solid by its own bounding
box)~\cite{kolodiazhnyi2026cadrille,elistratov2026cadevolve},
oriented surface coverage (OSC) and feature retention (FeatR95) at
$\tau=10^{-3}$ (Sec.~\ref{sec:boundary}). On CADBench we use its own
scorer: IoU after its pose alignment, and surface IoU (SIoU). B-rep search is evaluated at a shared
budget of $V=100$ kernel evaluations unless stated otherwise.

\subsection{Programs from different observations}
\label{sec:exp_cross}
\begin{table*}[!t]
\centering\small
\setlength{\tabcolsep}{3.6pt}
\begin{tabular}{@{}llrrrrrrrrrr@{}}
\toprule
& & & \multicolumn{6}{c}{748 parts} & \multicolumn{3}{c}{CADBench 348}\\
\cmidrule(lr){4-9}\cmidrule(lr){10-12}
Input & Method & Params & IND & $\geq.95$ & OSC & FeatR95 & Verified & Exact & IoU & SIoU & Time\\
\midrule
points & CAD-Recode v1.5~\cite{rukhovich2025cadrecode} & 1.54B & .482 & .110 & .188 & .050 & \textbf{7} & \textbf{3} & .544 & .471 & \\
points & cadrille-RL~\cite{kolodiazhnyi2026cadrille} & 2.21B & \textbf{.641} & .134 & .208 & \textbf{.064} & 6 & 1 & \textbf{.689} & \textbf{.583} & \\
\ours points & \sysname & 17.2M & .609 & \textbf{.250} & \textbf{.320} & .058 & 0 & 0 (260$^\dagger$) & .654 & .561 & \\
\midrule
mesh & CADENA~\cite{kabisov2026cadena} & 2.21B & .672 & .332 & .392 & .173 & 18 & 4 & .703 & .606 & \\
mesh & StepCAD~\cite{nehme2026stepcad} & 1.55B$^\ast$ & .546 & .136 & .213 & .085 & 11 & 4 & .663 & .548 & 177\\
mesh & CADFit~\cite{nehme2026cadfit} & 0.3B$^\ast$ & .444 & .302 & .445 & .326 & 66 & 0 & .545 & .469 & 1,804\\
\ours mesh & \sysname, search only & 17.2M & .648 & .594 & .672 & .607 & 273 & 99 & .820 & .696 & 338\\
\ours mesh & \sysname & 17.2M & \textbf{.864} & \textbf{.814} & \textbf{.793} & \textbf{.731} & \textbf{274} & \textbf{102} (539$^\dagger$) & \textbf{.937} & \textbf{.794} & 338\\
\midrule
drawing & Ortho2CAD RL~\cite{joglekar2026ortho2cad} & 8.77B & .334 & .008 & .069 & .005 & \textbf{1} & 0 & .443 & .323 & \\
\ours drawing & \sysname & 1.67M & \textbf{.744} & \textbf{.349} & \textbf{.449} & \textbf{.132} & 0 & 0 & \textbf{.799} & \textbf{.662} & 68\\
\midrule
\ours B-rep & \sysname & 2.74M & .736 & .673 & .789 & .751 & 417 & 403 & .868 & .728 & 256\\
\bottomrule
\end{tabular}
\caption{\textbf{Programs from every input.} Each program is replayed
in the kernel and scored against the reference; failures score zero.
The 748 single-solid ABC parts are 400 ABC-EVAL files and 348 CADBench
parts; the last three columns use the 348, with CADBench's official
scorer for mean IoU and SIoU. Time: mean search or prediction seconds per part on the 348 parts,
for methods that search or optimize per part (for our mesh route,
surface reconstruction is excluded; end-to-end times are in the appendix) (ours on CPU workers of a
2$\times$ Intel Xeon 8352Y server, StepCAD on an NVIDIA L4 GPU with
12 vCPUs, CADFit on an NVIDIA RTX PRO 6000 GPU with 48 vCPUs). IND: normalized volume IoU (mean);
$\geq.95$: share of parts; OSC, FeatR95: Sec.~\ref{sec:boundary},
over each row's eligible parts (appendix); Verified: outputs passing the
boundary check (a separate test from Exact); Exact: programs with $|\mathrm{IoU}-1|\leq10^{-5}$ to the reference;
$^\dagger$programs that replay our reconstructed solid exactly (routes
that reconstruct a solid first).
\emph{Search only} omits the section-stack fallback. Point
baselines receive their native 256 points, ours 8,192 oriented points.
Params: learned parameters of the networks the scored route uses
($^\ast$backbone only). Bold: best per input
group. Shading marks ours.}
\label{tab:programs}
\end{table*}

Table~\ref{tab:programs} compares executable programs on the same 748
parts. From meshes, \sysname reaches mean IND 0.864, compared with 0.672
for the strongest competing program method, CADENA. Its oriented
surface coverage and feature retention reach 0.793 and 0.731,
respectively. Search alone reaches 0.648 (Table~\ref{tab:programs}, search only);
the section-stack fallback, chosen on 337 of the 748 parts, raises
this to 0.864 while still returning an
executable program. From observed B-reps, \sysname recovers 403 volume-exact
programs; its mesh route recovers 102, versus at most four from any
external baseline in Table~\ref{tab:programs}. These inputs provide different levels of geometric
information, so the B-rep row measures recovery with the boundary
already known. The B-rep route has no section-stack fallback, so parts it
cannot recover exactly lower its mean IND (0.736) below the mesh route's. From dense points, our programs reach 0.609 IND against
0.641 for cadrille-RL. They reach IND of at least 0.95 on nearly twice as many
parts (0.250 against 0.134) and cover more surface (OSC 0.320 against
0.208), with slightly lower FeatR95 (0.058 against 0.064).

For drawings, the recovered programs reach 0.744 mean IND against 0.334
for Ortho2CAD RL; neither method produces a reference-exact program.
Without the learned correction, the carve program alone reaches mean IND 0.615 on
the 400 ABC-EVAL files and 0.748 on the 348 CADBench parts, against 0.689
and 0.807 with it (appendix). Table~\ref{tab:programs} lists the learned
parameters behind each row.

\subsection{Exact program recovery from B-reps}
\label{sec:exp_brep}
\begin{table}[t]
\centering\small
\setlength{\tabcolsep}{5pt}
\begin{tabular}{@{}lrr@{}}
\toprule
Method & $V=20$ & $V=100$\\
\midrule
Gallery agent (published)$^\dagger$~\cite{willis2021fusion360} & 62.3 & 67.5\\
Trained policy & 67.5 & 74.4\\
Boolean search (no learned parameters) & 80.1 & 83.0\\
\ours Boolean search + trained policy & \textbf{81.6} & \textbf{86.5}\\
\bottomrule
\end{tabular}
\caption{\textbf{Program recovery from B-reps.} Volume-exact programs
(\%) on the 1,725 Fusion~360 Gallery test designs within $V$ kernel
evaluations. $^\dagger$Reported in its own environment, where $V$
counts environment steps. Further baselines and subsets are in the
appendix.}
\label{tab:brep_main}
\end{table}

\begin{table}[t]
\centering\footnotesize
\setlength{\tabcolsep}{2.6pt}
\begin{tabular}{@{}llcccccc@{}}
\toprule
& & \multicolumn{3}{c}{ABC Extrude} & \multicolumn{3}{c}{ABC All-Ops}\\
\cmidrule(lr){3-5}\cmidrule(lr){6-8}
Method & Input & low & med & high & low & med & high\\
\midrule
CADFit~\cite{nehme2026cadfit} & mesh & .973 & .835 & .835 & .934 & .789 & .789\\
CADEvolve~\cite{elistratov2026cadevolve} & mesh & .928 & .548 & .621 & .855 & .541 & .480\\
CAD-Recode~\cite{rukhovich2025cadrecode} & mesh & .917 & .424 & .540 & .837 & .485 & .415\\
cadrille~\cite{kolodiazhnyi2026cadrille} & mesh & .911 & .485 & .554 & .840 & .514 & .430\\
\ours \sysname & mesh & \textbf{.996} & \textbf{.983} & \textbf{.988} & \textbf{.993} & \textbf{.972} & \textbf{.964}\\
\ours \sysname, search only$^\dagger$ & mesh & .996 & .978 & .980 & .989 & .956 & .857\\
\ours \sysname, with fallback$^\dagger$ & mesh & .996 & .979 & .990 & .991 & .972 & .958\\
\midrule
\ours \sysname & STEP & 1.000 & 1.000 & .999 & .997 & .970 & .876\\
\ours \quad exact programs & STEP & \multicolumn{3}{c}{2,499} & \multicolumn{3}{c}{1,391}\\
\bottomrule
\end{tabular}
\caption{\textbf{CADBench.} Median aligned IoU per face-count tier
(1,000 designs each; failures score zero). Published rows are from
CADBench~\cite{doris2026cadbench}. Our first mesh row is the selected
solid; the search-only and with-fallback rows score the returned
\emph{programs}, without and with the section-stack fallback.
The STEP rows return programs, with the number of exact programs out of
3,000 per split. $^\dagger$On the 348-design CADBench sample (58 per tier).
DeepCAD and Fusion~360 are in the appendix. Shading marks ours.}
\label{tab:cadbench_main}
\end{table}

\paragraph{Fusion~360 Gallery.}
At $V=100$, \sysname recovers 1,492 of 1,725 test designs exactly
(86.5\%; Table~\ref{tab:brep_main}). Boolean search, with no learned
parameters, accounts for 1,431 recoveries; the trained policy adds
61 more. The published Gallery agent reports 67.5\% at 100 steps in its
own environment~\cite{willis2021fusion360}. Our improvement is also
visible at $V=20$ (81.6\% versus 62.3\%, or 64.5\% with augmented training).

\paragraph{CADBench.}
Across the four 3,000-design splits, B-rep search finds 2,930 exact
programs on DeepCAD, 1,968 on Fusion~360, 2,499 on ABC Extrude and
1,391 on ABC All-Ops (ABC splits in Table~\ref{tab:cadbench_main}). The decline on
ABC All-Ops exposes the limits of the supported construction tools.
Scored as solids rather than programs, the mesh route's selected solids
reach median IoU 0.988 on the hardest ABC Extrude tier and 0.964 on ABC
All-Ops, against 0.835 and 0.789 for CADFit's programs. The mesh and
STEP rows use different inputs and output types.

\subsection{Search analysis and boundary fidelity}
\label{sec:exp_analysis}
\label{sec:code_evidence}
\label{sec:exp_features}

\paragraph{Expert iteration and search complementarity.}
Run alone at $V=24$, the trained policy recovers 174 of the 533
ABC-EVAL files exactly, against 155 for the imitation policy (paired exact test, $p=0.000157$). On Fusion~360 Gallery designs, the median conditional code length of
programs found by the trained policy decreases from 1.76 bits under the
imitation policy to 0.09 bits under the trained policy, while programs
found by Boolean search cost 7.58 bits under the trained policy.
The two searches therefore find different programs.
Component ablations and search-cost analyses appear in the appendix.

\begin{figure*}[!t]
\centering
\includegraphics[width=\textwidth]{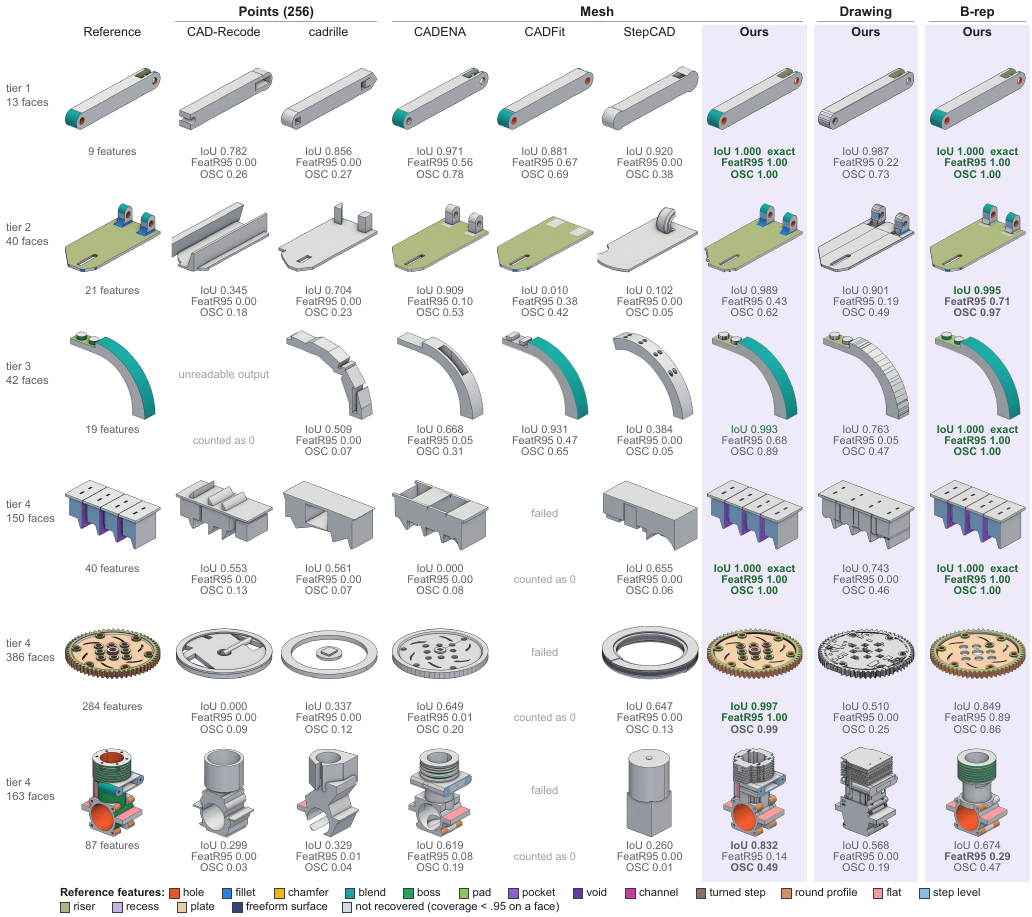}
\caption{\textbf{Programs from every input on single-solid ABC parts}
across face-count tiers; columns follow Table~\ref{tab:programs}.
Reference faces are colored by recognized feature, and an output keeps
a feature's color where every face of that feature is covered. Below
each output are its IND (labeled IoU), FeatR95 and OSC (best per row in bold);
\emph{exact} marks volume-exact programs, and failed outputs count as
zero.}
\label{fig:qual_abc}
\end{figure*}

\paragraph{Boundary fidelity.}
The distinction between volume fit and surface preservation is
visible in Table~\ref{tab:programs}: cadrille reaches 0.641 IND but
only 0.208 OSC and 0.064 FeatR95. On one part, removing a 0.2\,mm chamfer leaves volume IoU at
0.99994 but lowers OSC to 0.704, since OSC weights small patches as
much as large ones. Figure~\ref{fig:qual_abc}
illustrates which source features survive different reconstruction
routes.

\section{Limitations}
\label{sec:limits}

The recovery of exact construction programs depends on the
operations supported by our tool vocabulary. While target-derived
extrusions, revolutions and finishing operations cover many
mechanical parts, geometries requiring unsupported freeform
constructions remain challenging. Furthermore, a program that
reproduces the target geometry need not recover the original
modeling history or its design constraints.

For meshes, point clouds and drawings, reconstruction quality
also depends on how accurately the input geometry can be
recovered. Section-stack programs provide an executable fallback
when more compact constructions cannot be found, although
their sequences of extrusions may be less intuitive for subsequent
editing. Finally, boundary and feature-level measurements depend
on geometric tolerances and, for recognized features, the source
surface representation.

\section{Conclusion}
\label{sec:conclusion}

We presented \sysname, a geometry-guided framework for recovering
executable CAD programs from different geometric observations.
Our central insight is that an observed B-rep already contains the
geometry needed to define candidate construction operations,
reducing program recovery to a search over how those operations
are selected and combined. By coupling efficient Boolean search
with a compact learned policy, \sysname recovers construction
programs from B-reps without predicting their numerical parameters. The
framework also extends to meshes, point clouds and orthographic
drawings through geometric reconstruction, while boundary- and
feature-level evaluation captures details that volumetric
agreement alone can overlook.

More broadly, this approach opens a promising direction for CAD
automation: combining retrieval from domain-specific STEP libraries
with program recovery to transform existing geometric assets into
reusable construction programs. Such systems could exploit the
precision and diversity of available engineering models without
requiring large generative models to reconstruct every design
from scratch.
{
    \small
    \bibliographystyle{ieeenat_fullname}
    \bibliography{main}
}

\clearpage
\appendix
\section*{Appendix}
This appendix gives further analysis of the main results and details
of the experiments summarized in the main paper. It is organized as
follows. Section~\ref{sec:supp_reading} discusses the main results by
route and by learned component. Section~\ref{sec:supp_brep} covers
program recovery from B-reps: the action families, Boolean search, the
policy and its training, recovery on the Fusion~360 Gallery and
ABC-EVAL, wall time and model size, stand-alone CadQuery programs, and
ablations. Sections~\ref{sec:supp_cadbench}, \ref{sec:supp_mesh},
and~\ref{sec:supp_drawings} report the CADBench, mesh-and-point, and
drawing experiments. Section~\ref{sec:supp_metrics} defines the boundary
metrics, and Sec.~\ref{sec:supp_abc} describes the ABC-EVAL protocol and
the external baselines. Section~\ref{sec:supp_related_more} discusses
additional related work, and Sec.~\ref{sec:supp_featstrat} breaks down
the results by part complexity. Finally, Sec.~\ref{sec:supp_exec}
describes the entity representation and program execution, and
Sec.~\ref{sec:supp_mdl} gives description-length diagnostics. References
to tables, figures, and sections of the main paper say so explicitly.

\paragraph{Terms.}
In addition to the terms in the main paper, we use the following. The
\emph{Carrier route} assembles a solid from fitted and predicted
carriers (the analytic surfaces underlying faces, not trimmed to face
boundaries). A
\emph{section stack} fuses extrusions of profiles fitted to successive
mesh slices. The \emph{Full pool} chooses among all candidates by their
fit to the input mesh, whereas the \emph{Restricted pool} does the same
without section stacks and patch candidates. The \emph{priority
selector} instead takes the first usable candidate, starting with the
Carrier route. A \emph{patch candidate} corrects a candidate solid using
prisms over the pieces where it differs from the input mesh. Each prism
is added or cut, and it is kept only if it improves the fit to the input
mesh.
The mesh route in the main paper selects its solid using the Full pool
and returns the shorter of two programs (Sec.~\ref{sec:supp_mesh}, Mesh
programs). The \emph{imitation policy} imitates synthetic programs,
while the \emph{trained policy} is obtained after 30 rounds of expert
iteration. A \emph{kernel evaluation} replays a new program prefix in
the CAD kernel. A program is \emph{volume-exact} when its fresh replay
satisfies $|\IoU-1|\leq10^{-5}$ against the reference. Unless stated
otherwise, volume-exactness is measured against the reference; the
$\dagger$ mark in Table~1 of the main paper instead counts programs
whose replay reproduces the route's own reconstructed solid exactly,
rather than the reference. IND measures volume IoU after centering each
solid on its own bounding box and scaling it uniformly by the inverse of
its largest side; failures score zero throughout. IND therefore ignores
errors in overall position and size, but not in proportions.
\emph{Training-side} and \emph{training-validation} parts are drawn from
our training sources and are not test parts.

\section{Further analysis of the main results}
\label{sec:supp_reading}

This section examines main-paper Tables~1 and~2 in more detail by route
and by learned component. Later sections give the details of each
experiment.

\paragraph{Mesh route: search programs and section stacks.}
Without section stacks, mesh search alone reaches a mean IND of .648 on
the 748 parts, compared to .672 for CADENA. The paired bootstrap 95\%
interval of the difference, $-.023$ $[-.058,+.011]$, includes zero.
Search alone scores higher on every other metric in main-paper Table~1:
the share of parts with IND of at least .95 is .594 versus .332, OSC is
.672 versus .392, FeatR95 is .607 versus .173, Verified is 273 versus
18, and Exact is 99 versus 4. On the 348 CADBench parts, IoU and SIoU
are .820 and .696 versus .703 and .606. Search
alone produces no valid output on 182 parts compared to 131 for
CADENA; over valid outputs alone, the means are .857 and .814.
Split by subset, the difference is $-.134$ $[-.178,-.090]$ on the 400
ABC-EVAL files and $+.104$ $[+.052,+.156]$ on the 348 CADBench parts.
Adding the section-stack fallback returns a section-stack program for
337 parts and a search program for 360. The remaining 51 parts have no
output, 47 of them because they lack an input mesh. Mean IND rises from
.648 to .864, and Exact increases from 99 to 102. Section stacks are
executable CadQuery programs made of many slabs.
The returned search programs reach a mean IND of .968 with a median of
1 action and 34 output faces, whereas the returned section stacks reach
.884 with a median of 10 features and 224 faces. Written as scripts,
section stacks use a median of 101 CadQuery operations, compared to 3
operations and 23.5 output faces for CADENA's programs on the ABC-EVAL
parts (Table~\ref{tab:supp_compact}).
When we score the route on the solids built by stand-alone scripts of
its search programs rather than our interpreter's replay, mean IND is
.839 instead of .864, and the share of parts with IND of at least .95
is .789 instead of .814 (Table~\ref{tab:supp_script_scores}). We give
the selection rule in Sec.~\ref{sec:supp_mesh} (Mesh programs). In the
named-edit test, section stacks pass more often than search programs
(77.5\% against 67.5\%); when the same edit is applied on the same
parts, neither leads significantly ($p=.22$ and $.18$). The route is built for clean meshes: under 0.5\% vertex jitter
on 100 parts, its mean IND falls from .954 to .537, below cadrille-RL
and CADENA (.667 each). Under CADBench's milder published noisy-mesh
protocol, it keeps a mean CADBench IoU of .806 and a median of .923 on
96 parts, against a clean mean of .944 (Sec.~\ref{sec:supp_mesh}). See Tables~\ref{tab:supp_validonly},
\ref{tab:supp_selection}, \ref{tab:supp_compact},
\ref{tab:supp_named_edit} and~\ref{tab:supp_noise}.

\paragraph{Fusion~360 Gallery and the Gallery agent.}
Main-paper Table~2 places our 86.5\% next to the 67.5\% reported by the
Gallery agent in its own environment, which differs from ours in kernel,
exactness test, and budget unit. When we run the agent with its released
weights in our runtime, under our counter, and on the same 1,725
designs, it recovers 998 designs (57.9\%) at $V=100$, compared to 1,492
(86.5\%) for our full solver. Our solver recovers 505 designs that the
agent misses, whereas the agent recovers 11 that our solver misses. This
gain comes from pairing the search with a larger target-derived tool
set. Restricted to face extrusions, the action family the agent uses,
Boolean search recovers 890 designs. Adding revolves raises this count
to 1,273, and the full tool set reaches 1,431. Because $V$ counts only
replays of new program prefixes, we also compare wall time on the same
machine with all computation included. Within 60\,s, our solver recovers
1,444 designs compared to 912 for the agent, and the median time per
design is 11.4\,s versus 26.7\,s. Under the same wall-time limits,
Boolean search with face extrusions alone recovers nearly as many
designs as the agent within 60\,s (887 versus 912) and fewer within
600\,s (888 versus 998). On the 938 designs whose designer sequence has
at least two extrusions, it recovers more than the agent at both limits.
Sections~\ref{sec:supp_brep_results} and~\ref{sec:supp_time} report the
paired tests, the kernel operations behind $V$, a live run of the full
solver, and the agent's results on the 748 parts. See
Tables~\ref{tab:supp_brep_full}, \ref{tab:supp_families},
\ref{tab:supp_time_gallery} and~\ref{tab:supp_time_x}.

\paragraph{Contribution of the learned components.}
Adding the trained policy to Boolean search recovers 61 more Gallery
designs at $V=100$ (paired bootstrap 95\% interval $[+46,+76]$). On the
748 parts, the full solver recovers 403 compared to 400 for Boolean
search alone. When run alone on the 533 ABC-EVAL files, the trained
policy helps most at small budgets, recovering 88 files versus 67 for
the imitation policy at $V=1$, 174 versus 155 at $V=24$ ($p=.00016$),
and 181 versus 173 at $V=100$ ($p=.077$). The drawing correction network
raises mean IND over the carve program from .615 to .689 on the 400
ABC-EVAL files and from .748 to .807 on the 348 CADBench parts, while
OSC and FeatR95 barely change. On the Ortho2CAD test sets, it raises the
scores from .885 and .823 to .893 and .831. Across sizes from 2.74M to
125M parameters, B-rep policies recover 122 to 138 of 293 training-side
parts at $V=100$, and three seeds of the smallest model alone span 122
to 135. Section~\ref{sec:supp_time} lists the networks behind each row
of main-paper Table~1 along with their sizes and run times. See
Tables~\ref{tab:supp_brep_full}, \ref{tab:supp_families},
\ref{tab:supp_budget}, \ref{tab:supp_drawings}
and~\ref{tab:supp_policy_size}, and
Secs.~\ref{sec:supp_brep_results} and~\ref{sec:supp_drawings}.

\paragraph{Inputs of the point and drawing comparisons.}
Our point route takes 8,192 oriented points, whereas the point baselines
take their native 256. Given 256 points, our route reaches .556 IND
compared to .641 for cadrille-RL. For drawings, Ortho2CAD takes sheets
from its own released generator, which prints bounding-box dimensions.
Our route takes sheets in the same layout without dimension text. When
evaluated on sheets from Ortho2CAD's generator across all 748 parts, our
route scores .725 IND compared to .334 for Ortho2CAD RL, a paired
difference of $+.39$ $[+.37,+.41]$. On the drawings from Ortho2CAD's own
test sets, our route scores .893 on DeepCAD and .831 on Fusion~360,
compared to .776 and .560 for Ortho2CAD RL. See
Tables~\ref{tab:supp_points748}, \ref{tab:supp_o2csheets}
and~\ref{tab:supp_drawings}, and Sec.~\ref{sec:supp_drawings}.

\begin{table}[t]
\centering\footnotesize
\setlength{\tabcolsep}{2.5pt}
\begin{tabular}{@{}lrrrrrr@{}}
\toprule
Method and selection & IND & $\geq.95$ & OSC & FeatR95 & Exact & Fail.\\
\midrule
\multicolumn{7}{@{}l}{\emph{All 748 parts, 256 input points}}\\
CAD-Recode v1.5, greedy & .482 & .110 & .188 & .050 & 3 & 212\\
\quad best of 8, fit to its points & .611 & .124 & .207 & .055 & 2 & 89\\
\quad best of 8, by reference IND & .646 & .139 & .213 & .061 & 2 & 62\\
cadrille-RL, greedy & .641 & .134 & .208 & .064 & 1 & 59\\
\quad best of 8, fit to its points & .662 & .146 & .210 & .064 & 1 & 57\\
\quad best of 8, by reference IND & .683 & .158 & .219 & .070 & 1 & 57\\
\midrule
\multicolumn{7}{@{}l}{\emph{96 of the 748 parts, mesh input}}\\
CADENA, one rollout & .692 & .323 & .427 & .202 & 0 & 18\\
\quad best of 4, own reward & .738 & .365 & .450 & .232 & 0 & 14\\
\midrule
\ours Ours, 8,192 points, 748 parts & .609 & .250 & .320 & .058 & 0 & 155\\
\ours Ours, mesh, 748 parts & \textbf{.864} & \textbf{.814} & \textbf{.793} & \textbf{.731} & \textbf{102} & 64\\
\bottomrule
\end{tabular}
\caption{\textbf{Sampling more programs from the baselines.} Greedy and
one-rollout rows repeat main-paper Table~1 on the same parts. Best of $N$ keeps one of
$N$ sampled programs, chosen by fit to the method's own input (for
CADENA, its own reward, the fit to its input mesh) or, as an upper
bound for the baselines, by IND against the reference. On clean meshes,
which are tessellations of the reference, that bound uses about the same
information as our mesh selector, which ranks candidates by fit to the
input mesh. The CADENA rows cover 96 parts, 48 from ABC-EVAL and 48 from
CADBench, stratified by tier. Failures score zero; Fail.\ counts parts
without a scorable output.}
\label{tab:supp_bestofn}
\end{table}

\paragraph{Sampling more programs from the baselines.}
Our routes search for minutes per part, whereas CAD-Recode v1.5 and
cadrille-RL decode one program in about 10\,s and CADENA runs one
rollout in about 80\,s.
Table~\ref{tab:supp_bestofn} gives them more samples. CAD-Recode v1.5
and cadrille-RL sample eight programs per part on all 748 parts, and
CADENA samples four rollouts on 96 of them. Sampling helps every
baseline, CAD-Recode most: its failures fall from 212 to 89 and its
mean IND rises from .482 to .611 ($+.129$ $[+.109,+.149]$).
cadrille-RL rises from .641 to .662 ($+.021$ $[+.015,+.028]$), which
widens its lead over our point route (.609), and CADENA rises from .692
to .738 on its 96 parts ($+.046$ $[+.017,+.082]$). Choosing by IND against
the reference, an upper bound, raises CAD-Recode and cadrille-RL to
.646 and .683. No
baseline gains an exact program, and OSC rises by at most .025. On all
748 parts, sampling leaves CAD-Recode and cadrille-RL well below our
mesh route (.864), and CADENA's gain on its 96 parts ($+.046$) is small
next to its gap to our mesh route in main-paper Table~1 ($.192$).

\paragraph{Returned programs and stand-alone scripts.}
Exact means volume-exact against the reference. Verified is a separate
test: the output passes the bidirectional boundary check at
$\tau=10^{-3}$, whether or not it is exact. Programs from the B-rep
route are action lists on the target's faces, which an emitter writes as
stand-alone CadQuery scripts. Section-stack and drawing programs are
CadQuery scripts by construction. On test parts, the emitted scripts
reproduce the program's solid within $10^{-5}$ for 97.8\% of the Gallery
programs and 91.1\% of the programs on the 748 parts. When scored on the
solids built by these scripts, the full solver recovers 1,471 of the
1,725 Gallery designs at $V=100$ instead of 1,492, the B-rep row of
main-paper Table~1 reaches .681 IND instead of .736, and the mesh row
reaches .839 instead of .864.
The B-rep counts on the 748 parts measure different quantities.
Exact and the emitter check count programs returned within $V=100$;
the other columns of the row score the program that each part returns
within its wall limit, which can come after $V=100$.
Main-paper Table~1 counts 403 programs that are volume-exact within
$V=100$ under our interpreter's replay. Boolean search alone reaches 400
within $V=100$ and 424 with no limit on $V$
(Table~\ref{tab:supp_families}). With no limit on $V$, the stand-alone
scripts for the row's outputs are volume-exact on 417 parts (see the
text after Table~\ref{tab:supp_script_scores}). The emitter check evaluates the
560 programs returned within $V=100$: 529 are written as scripts, and
393 produce a volume-exact solid (Tables~\ref{tab:supp_cqemit}
and~\ref{tab:supp_edit}). The row returns a program for 674 parts, of
which 668 build a solid; 605 of these give a valid, scorable output, so
the row records 143 failures (Table~\ref{tab:supp_validonly}). In the named-edit test, 67.5\% of our
B-rep search programs, 77.5\% of our section stacks, and 68.5\% of
CAD-Recode v1.5 programs pass all three checks. When the same edit,
defined from the reference part, is applied on 150 parts where all three
return a program, they pass on 38\%, 33\% and 25\% (search programs
against CAD-Recode, $p=.010$); on 150 parts drawn from all 748, they
pass on 29\%, 35\% and 21\%. See
Tables~\ref{tab:supp_verified}, \ref{tab:supp_cqemit},
\ref{tab:supp_script_scores}, \ref{tab:supp_edit},
\ref{tab:supp_named_edit}, \ref{tab:supp_paired_edit}
and~\ref{tab:supp_compact}.

\section{Program recovery from B-reps}
\label{sec:supp_brep}

\begin{table*}[t]
\centering\small
\setlength{\tabcolsep}{4pt}
\begin{tabular}{@{}p{.15\textwidth}p{.47\textwidth}p{.19\textwidth}p{.12\textwidth}@{}}
\toprule
Family & Operand and its geometric source in the target $T$ & Possible effects & Used by\\
\midrule
Extrude (X) & Prism of a planar target face, outer and inner loops included, up to another target plane: the exact distance if the planes are parallel, otherwise a long prism clipped by that plane's half-space & new body, join, cut, intersect & both solvers\\
Extrude from the defeatured solid (XD) & As X, but the profile is the face of the defeatured solid $T_d$ that is coplanar with the target face and contains its center & new body, join, cut, intersect & both solvers\\
Hole (H) & Cylinder from a concave cylindrical face flagged as a hole (blind holes included): axis, radius and axial extent of the face & cut & both solvers$^\ast$\\
Revolve (R) & Side 0: the section of $T_d$ by a half-plane through the face's axis, revolved; side 1: the envelope on the cut side & join or new body (side 0), cut (side 1) & both solvers\\
Fillet (F) & Edges of the current solid between the blend's two support surfaces; radius from the blend face & finishes the current solid & both solvers$^\ast$\\
Chamfer (C) & Chamfer distances from the target chamfer face & finishes the current solid & both solvers$^\ast$\\
Shell (S) & Removes the rim face; wall thickness from the distance between the rim face's two wires & finishes the current solid & both solvers$^\ast$\\
Pattern (P) & Repeats the previous action's tool by a symmetry of $T$ (rotation, reflection or translation), found by a Boolean common of $T$ with its transformed copy & repeat & policy only\\
\bottomrule
\end{tabular}
\caption{\textbf{B-rep action families.} Every operand is read or derived
from faces of the target $T$ (the first STEP root); no network predicts a
numeric value. For B-spline blend faces, the fillet radius comes from a
constant-curvature estimate on the face. The defeatured solid $T_d$ is
produced from $T$ by removing blend, hole, and chamfer faces with Open
CASCADE defeaturing. The first action must create a new body, and pattern
is not applied after F, C, or S. The action set does not include loft,
draft, or extrusion along a curved path. $^\ast$In Boolean search, F, C,
and S are used only in the finishing stage, while H is used in both the
body and finishing stages (Sec.~\ref{sec:supp_search}); Boolean search
never uses P.}
\label{tab:supp_actions}
\end{table*}

\paragraph{Actions.}
Table~\ref{tab:supp_actions} lists the action families and the
geometric source of each operand. Every action names faces of $T$ by
their index in the kernel's face map. X extrudes a target face directly,
bringing its outer and inner loops along with it. XD instead takes its
profile from the defeatured solid $T_d$, allowing a profile interrupted
by blends, holes, or chamfers to be extruded whole while later actions
restore the removed features. Together, X and XD form the extrusion
family in the main paper. Revolve sections are also cut from $T_d$. The
policy uses all eight families, while Boolean search uses X, XD, H, and
R for the body and F, C, S, and H for finishing. Face extrusions and
revolves follow prior
work~\cite{willis2021fusion360,xu2021zonegraph,yin2025rlcad}; holes,
fillets, chamfers, shells, and defeatured bodies extend this tool set.
Table~\ref{tab:supp_families} adds these families to Boolean search one
at a time.

\subsection{Boolean search}
\label{sec:supp_search}

\paragraph{Probes and bits.}
The probe set combines uniform volume probes with face-pair probes. We
draw 1,024 points uniformly in the bounding box of $T$, padded by 2\% of
its diagonal, using a per-part seed. For each target face, we also take
up to three interior points $p$ from a $10\times10$ grid inside its
trim, yielding $p-\epsilon n$ (inside or on $T$) and $p+\epsilon n$
(outside), where $\epsilon=2\cdot10^{-4}$ of the diagonal. The kernel
classifies every probe against $T$. We compute tool bits analytically
for X and XD using a two-dimensional trim test of the profile plus a
height or half-space test, and with the kernel classifier for H and R.
We drop tools with no inside bit, merge tools with identical bits, and
keep at most 20,000 tools.

\paragraph{Beam search and refinement.}
Search starts from roots formed by a single X, XD, or R tool as a new
body. At each depth, we apply every tool to every state with join (OR),
cut (AND-NOT), or intersect (AND), and merge states with identical bits.
The distance to $T$ is the popcount of the XOR between bit vectors, with
face-pair and refinement bits weighted $4\times$. The beam keeps 512
states up to depth 16 and returns up to ten zero-distance programs,
which the kernel replays in turn. If a replay is not volume-exact, we
add three probes in each of the ten largest pieces of the kernel
differences $T\setminus U$ and $U\setminus T$ and restart the beam, up
to five refinements. Search stops at the first volume-exact program
whose solid places every sampled target face on its boundary; otherwise
it returns the first volume-exact program, or none if no replay is
volume-exact. Probe agreement is exact Boolean algebra on the chosen
points, while kernel replay decides whether a program rebuilds $T$.

\paragraph{Features.}
When $T$ has blends, holes, or chamfers, the body stage can target
$T_d$. Finishing actions (F, C, S, and H) are then searched over action
sets with a beam of width 3, capped at 160 kernel calls.

\paragraph{Budget accounting.}
Each kernel evaluation replays a new program prefix, so later programs
that share an existing prefix pay only for their new prefixes. Wall time
includes all stages, whereas the evaluation counter does not charge for
probe scoring, tool construction, defeaturing, probe classification,
counterexample differences, the face test, symmetry detection, policy
featurization, or network calls. Policy search stops when it reaches $V$
evaluations. Boolean search runs to its wall limit and then applies $V$
to its recorded count; a missing record is charged $V$, and the policy
receives the remaining budget. We run the released Gallery agent in our
runtime under the same counter. On average, Boolean search takes
213.2\,s per part on ABC-EVAL and 185.4\,s on CADBench ABC All-Ops.

\subsection{Policy and expert iteration}
\label{sec:supp_policy}

\paragraph{Network.}
The policy is a 2.74M-parameter graph transformer (2,741,908 parameters;
width 168, four layers, six heads). Each target face has 740 input
features: its surface type, 100 sampled positions and normals on a UV
grid, 100 inside/on bits, and 32 descriptors (axis, offset, radius,
area, perimeter, edge counts by convexity, and hole, blend, chamfer, and
rim flags). Faces connect through shared edges, carrying shortest-path
hop buckets and eight attributes per edge. After every step, we encode
the current solid with the same features. Parameter-free comparisons of
target and current face samples (coverage, distances, and inside/outside
shares) enter per target face, per current face, and per pair. A shared
encoder combines edge-conditioned message passing with attention biased
by adjacency, hops, and edge attributes, and cross-attention lets target
faces read the current solid. Output heads predict the action family
first and then its arguments: start face, end face conditioned on the
start face, and operation for X and XD; one face for H, F, C, and S; a
face and side for R; and a symmetry for P. Each family's argument
probabilities are normalized over its valid actions. The network has no
value head.

\paragraph{Imitation and expert iteration.}
We train the imitation policy on Zero-to-CAD~\cite{ataei2026zerotocad},
a set of synthetic CAD programs generated by language-model agents. Each
round of expert iteration~\cite{anthony2017exit} draws 192 fresh
Zero-to-CAD training targets without reuse and runs one best-first
search per target with the current policy, capped at 24 kernel
evaluations and 120\,s. We accept a program when search reaches a
volume-exact solid ($10^{-5}$), followed by a replay check at IoU
$\geq.999$. On the accepted states, the update minimizes the negative
log-likelihood of the accepted actions plus $0.02$ times the exact KL
divergence to the imitation policy over valid actions, with unit weight
on every accepted step. Each round trains for two epochs of eight
minibatches with AdamW (learning rate $3\cdot10^{-5}$, gradient clip 1).
During rounds 21--30, search solves about 46\% of its targets (roughly
89 accepted programs per round), while the per-state KL grows from .59
in rounds 1--10 to 1.48 in rounds 21--30. We use the policy after 30
rounds. During search, we pool sibling actions that build the same tool
with the same operation and avoid expanding states already reached. The
near-uniform control runs the same search using the trained policy at
temperature 100.

\subsection{Recovery on the Fusion~360 Gallery and ABC-EVAL}
\label{sec:supp_brep_results}

\begin{table*}[t]
\centering\small
\setlength{\tabcolsep}{5pt}
\begin{tabular}{@{}lrrrrrrrr@{}}
\toprule
& \multicolumn{3}{c}{Gallery release subset (982)}
& \multicolumn{2}{c}{Gallery full split (1,725)}
& \multicolumn{3}{c}{ABC single solids (400)}\\
\cmidrule(lr){2-4}\cmidrule(lr){5-6}\cmidrule(l){7-9}
Method & $V=1$ & $V=100$ & No 1-op & $V=20$ & $V=100$ & $V=24$ & $V=100$ & Verified$_{100}$\\
\midrule
Gallery agent (published)$^\dagger$~\cite{willis2021fusion360} & -- & -- & -- & 62.3\% & 67.5\% & -- & -- & --\\
Gallery agent (released, our runtime) & \textbf{681} & 783 & -- & 902 & 998 & -- & 81 & --\\
Near-uniform policy & 374 & 756 & -- & -- & -- & -- & -- & --\\
Imitation policy & 472 & 798 & 117 & -- & -- & 104 & 121 & 117\\
Trained policy & 638 & \textbf{869} & \textbf{130} & 1,165 & 1,283 & \textbf{120} & 127 & \textbf{123}\\
Boolean search & -- & -- & -- & 1,382 & 1,431 & -- & -- & --\\
\ours Boolean search + trained policy & -- & -- & -- & \textbf{1,408} & \textbf{1,492} & -- & \textbf{171} & --\\
\ours \quad live run of the full solver & -- & -- & -- & 1,406 & 1,490 & -- & -- & --\\
\bottomrule
\end{tabular}
\caption{\textbf{Program recovery from B-reps, all solvers and subsets.}
Volume-exact programs ($|\IoU-1|\leq10^{-5}$ on fresh kernel replay)
found within $V$ kernel evaluations. Under one shared budget, the full
solver runs Boolean search first and passes any unused budget to the
trained policy. ``Live run of the full solver'' runs this combination
directly; the other combined counts are assembled from standalone traces
of the two solvers. The released Gallery agent runs in our runtime with
its released weights, on the same designs and under our counter; it uses
face extrusions only. No 1-op reports recovery on the 217 release
designs for which none of our solvers finds a one-operation solution.
The ABC columns cover the 400 single-solid files from the 533-file
ABC-EVAL set (first STEP root); Verified$_{100}$ requires both volume
exactness and the boundary check against the whole file.
$^\dagger$Results from Willis et al.~\cite{willis2021fusion360} (Table~3)
in their own environment, where the budget counts environment steps in
the Fusion~360 Gym; with augmented training, the agent reaches 64.5\% at
20 steps (their Table~4). --: not run or not reported.}
\label{tab:supp_brep_full}
\end{table*}

\paragraph{Gallery subsets.}
The 982 release designs are the official test designs with a released
extrude-only sequence. The 1,405 designs pass the official static
filter, and the remaining 320 complete the 1,725 test designs
($982\subset1{,}405\subset1{,}725$). Because we reconstructed the 982
and 1,405 subsets ourselves, they differ from the 1,412-design
extrude-only list of Willis et al.; only the 1,725 split shares the same
denominator as the published agent. On the 1,405 designs, combined
search recovers 1,219 at $V=20$ and 1,283 at $V=100$, compared with
1,198 and 1,234 for Boolean search alone; on the 320 designs, combined
search recovers 189 and 209 against 184 and 197. On all 1,725 designs at
$V=100$, the trained policy adds 61 recoveries to Boolean search
(Table~\ref{tab:supp_brep_full}), even though its training data contains
no Fusion~360 design. This 61-design gain at $V=100$ has a paired
bootstrap 95\% interval of $[+46,+76]$ over 10,000 resamples, with all
61 discordant designs favoring combined search (exact McNemar test,
$p\approx9\times10^{-19}$); at $V=20$, the gain is $+26$ $[+17,+37]$.
When we run the released Gallery agent with its released weights in our
runtime under our counter on the same 1,725 designs, it recovers 902
designs at $V=20$ and 998 at $V=100$ (52.3\% and 57.9\%), compared with
1,408 and 1,492 for the full solver. Its published rates in its own
environment are 62.3\% and 67.5\%. Because the two settings differ in
kernel, exactness test, and budget unit, only the same-runtime rates
compare directly with ours. At $V=100$, the full solver uniquely
recovers 505 designs while the agent uniquely recovers 11. Because the
agent uses face extrusions alone, it recovers 783 of the 982 release
designs but only 80 of the 320 designs outside the static filter. When
we count a design only if the solid of its stand-alone script is
volume-exact to the input, the full solver recovers 1,391 designs at
$V=20$ and 1,471 at $V=100$ (Table~\ref{tab:supp_script_scores}).

\paragraph{ABC-EVAL, all 533 files.}
At $V=24$, the budget used in each expert-iteration search
(Sec.~\ref{sec:supp_policy}), the imitation policy recovers 155
volume-exact programs and the trained policy recovers 174 (152 by both,
22 only by the trained policy, and three only by the imitation policy).
At $V=100$, those counts are 173 and 181.
Table~\ref{tab:supp_budget} reports the full budget curve for policy
search alone. The trained policy recovers more files at every budget,
with the widest margin at small budgets: 88 against 67 at $V=1$. Within
21 evaluations, it already matches the imitation policy's $V=100$ count
(173). Across the 169 files recovered by both policies at $V=100$, the
trained policy uses fewer evaluations on 63 and more on 14.

\begin{table}[t]
\centering\footnotesize
\setlength{\tabcolsep}{4pt}
\begin{tabular}{@{}lrrrr@{}}
\toprule
Kernel evaluations $V$ & 1 & 8 & 24 & 100\\
\midrule
\multicolumn{5}{@{}l}{\emph{All 533 files}}\\
Imitation policy & 67 & 138 & 155 & 173\\
\ours Trained policy & \textbf{88} & \textbf{157} & \textbf{174} & \textbf{181}\\
Gains / losses & 23 / 2 & 22 / 3 & 22 / 3 & 12 / 4\\
Paired exact test, $p$ & .00002 & .00016 & .00016 & .077\\
\midrule
\multicolumn{5}{@{}l}{\emph{400 single-solid files}}\\
Imitation policy & 34 & 89 & 104 & 121\\
\ours Trained policy & \textbf{50} & \textbf{105} & \textbf{120} & \textbf{127}\\
\bottomrule
\end{tabular}
\caption{\textbf{Policy search alone on ABC-EVAL, by budget.}
Volume-exact programs within $V$ kernel evaluations, with sibling
actions pooled as in the main paper. Gains / losses: files recovered
only by the trained policy / only by the imitation policy; the paired
exact test uses these files. The reference is the first STEP root
(Sec.~\ref{sec:supp_abc}).}
\label{tab:supp_budget}
\end{table}

\begin{table}[t]
\centering\footnotesize
\setlength{\tabcolsep}{2.5pt}
\begin{tabular}{@{}lrrrrrr@{}}
\toprule
Method & IND & $\geq.95$ & OSC & FeatR95 & Verified & Exact\\
\midrule
Gallery agent~\cite{willis2021fusion360} & .555 & .366 & .507 & .448 & 223 & 219\\
\ours \sysname & \textbf{.736} & \textbf{.673} & \textbf{.789} & \textbf{.751} & \textbf{417} & \textbf{403}\\
\bottomrule
\end{tabular}
\caption{\textbf{Program recovery from B-rep input on the 748 parts of
main-paper Table~1.} Both methods take the B-rep as input and are
evaluated in our runtime under the same kernel-evaluation counter at
$V=100$. The Gallery agent uses its released weights. Columns follow
main-paper Table~1, and failures score zero. Our row is the B-rep row
from main-paper Table~1.}
\label{tab:supp_brep748}
\end{table}

On the 748 parts from main-paper Table~1, we run the released Gallery
agent on the same B-rep inputs in our runtime under our counter. It
recovers 219 volume-exact programs with a mean IND of .555, compared
with 403 programs and .736 for our B-rep route
(Table~\ref{tab:supp_brep748}); our route uniquely recovers 185 parts
and the agent uniquely recovers one. On the 400 ABC-EVAL files, the
agent recovers 81 parts against 171 (Table~\ref{tab:supp_brep_full}).
Main-paper Sec.~4.2 compares the 102 Exact programs from our mesh route
with the point, mesh, and drawing baselines in main-paper Table~1, each
of which returns at most four. For B-rep input, the external method is
the Gallery agent, with the 219 programs reported above.

\subsection{Wall time and model size}
\label{sec:supp_time}
\begin{table}[t]
\centering\footnotesize
\setlength{\tabcolsep}{3pt}
\begin{tabular}{@{}lr@{}}
\toprule
\multicolumn{2}{@{}l}{\emph{Trained policy, release subset (982)}}\\
Solved within 1\,s & 0\\
Solved within 5\,s & 171\\
Solved within 10\,s & 675\\
Solved within 60\,s & 857\\
Solved within 300\,s & 869\\
Median seconds per design & 6.58\\
Mean kernel evals., $V=100$ & 9.99\\
\midrule
\multicolumn{2}{@{}l}{\emph{Trained policy, 320 designs outside the static filter}}\\
Median seconds per design & 47.53\\
\midrule
\multicolumn{2}{@{}l}{\emph{Boolean search, seconds per design (median / mean / P90)}}\\
1,405 static-filter designs & 1.7 / 16.7 / 29.2\\
320 remaining designs & 7.8 / 67.6 / 188.7\\
\midrule
\multicolumn{2}{@{}l}{\emph{All 1,725 test designs, ours (full solver) / Gallery agent}}\\
Solved within 1\,s & 0 / 0\\
Solved within 5\,s & 0 / 223\\
Solved within 10\,s & 542 / 476\\
Solved within 60\,s & 1,444 / 912\\
Solved within 300\,s & 1,484 / 995\\
Median seconds per design & 11.4 / 26.7\\
\bottomrule
\end{tabular}
\caption{\textbf{Search wall time on the Fusion~360 Gallery.} Each
design uses one CPU process. The upper blocks report the trained policy
alone; the imitation policy has a median of 4.94\,s on the release
subset. The middle block reports Boolean search. Wall time includes
uncharged work such as probe scoring, tool construction, and
defeaturing. The bottom block compares our full solver (Boolean search
followed by the trained policy) against the released Gallery agent on
the same machine, with all computation included in wall time. A design
counts as solved within a time limit if it is recovered at $V=100$ and
its total wall time, in seconds per design, is within that limit; the
median is computed over all 1,725 designs. Table~\ref{tab:supp_time_x}
reports seconds to the volume-exact program instead.}
\label{tab:supp_time_gallery}
\end{table}

\begin{table*}[t]
\centering\small
\setlength{\tabcolsep}{4pt}
\begin{tabular}{@{}llrrrrrl@{}}
\toprule
& & Learned & \multicolumn{2}{c}{348 parts} & \multicolumn{2}{c}{748 parts, end to end} & \\
\cmidrule(lr){4-5}\cmidrule(lr){6-7}
Method & Input & parameters & Mean s & Median s & Mean s & Median s & Hardware\\
\midrule
\ours \sysname & mesh & 17.2M & 337.8 & 95.7 & 1,455 & 1,299 & 2$\times$ Intel Xeon 8352Y\\
\ours \sysname & points & 17.2M & -- & -- & -- & -- & same\\
\ours \sysname & B-rep & 2.74M & 256.1 & 28.1 & 331.2 & 94.8 & same\\
\ours \sysname & drawing & 1.67M & 67.5 & 48.9 & -- & -- & same\\
StepCAD~\cite{nehme2026stepcad} & mesh & $\approx$1.55B$^\ast$ & 177.2 & 174.8 & 177.3 & 174.7 & NVIDIA L4, 12 vCPUs\\
CADFit~\cite{nehme2026cadfit} & mesh & $\approx$0.3B$^\ast$ & 1,804.0 & 526.9 & 1,859.7 & 916.5 & NVIDIA T4 and NVIDIA RTX PRO 6000\\
CADENA~\cite{kabisov2026cadena} & mesh & 2.21B & 87.7 & 63.4 & 77.8 & 56.0 & RTX 4090\\
CAD-Recode v1.5~\cite{rukhovich2025cadrecode} & points & 1.54B & 11.93 & 11.09 & 12.4 & 11.8 & RTX 4090\\
cadrille-RL~\cite{kolodiazhnyi2026cadrille} & points & 2.21B & 7.96 & 7.27 & 10.3 & 9.5 & RTX 4090\\
Ortho2CAD RL~\cite{joglekar2026ortho2cad} & drawing & 8.77B & -- & 26.62 & 24.7 & 23.7 & NVIDIA L4\\
\bottomrule
\end{tabular}
\caption{\textbf{Time and model size per method.} The 348-part columns
use the CADBench parts from Table~1 of the main paper. Their per-part
times cover different stages across methods. Our times are search
worker-seconds (for meshes, the search over the selected solid; the 400
ABC-EVAL parts take a mean of 544.2\,s), except that our drawing time
is prediction time. StepCAD covers model inference and search (318
parts returning code), CADFit covers per-part optimization wall time
(76 parts reached its 7,200\,s cap), and CADENA covers rollout time.
CAD-Recode and cadrille cover decoding, plus execution on the ABC-EVAL
half, and Ortho2CAD covers GPU generation (mean not logged). The
end-to-end columns give recorded seconds per part from input to
returned program. For our mesh route, they cover the 701 parts with an
input mesh and every stage before and after search; these numbers are
lower bounds because process start-up is not recorded. B-rep has no
reconstruction stage, so its end-to-end time is search and program
export. For external methods, the end-to-end columns report each
method's own recorded time per part over the parts with a record,
without scoring. CADFit ran on the NVIDIA RTX PRO 6000 for the 348
parts and on NVIDIA T4 and RTX PRO 6000 GPUs for the 400 ABC-EVAL
parts. StepCAD ran on NVIDIA L4 GPUs for the 348 parts; the GPU used
for its ABC-EVAL half was not recorded. External times are measured on
the GPUs named, and both CADFit and StepCAD require Linux. --: not
available. $^\ast$Backbone only: StepCAD's language model (its encoder
count is not available) and CADFit's DINOv2 ViT-L. Our learned
components are the point network (11.74M), completion network (11.83M),
candidate-stage face policy (2.75M), trained B-rep policy (2.74M), and
drawing correction network (1.67M). The mesh total of 17.2M counts the
point network, the candidate-stage face policy, and the trained policy.
Search, surface fitting, meshing, and program writers have no learned
parameters, and no route uses a language model.}
\label{tab:supp_time_params}
\end{table*}

On the Gallery, Boolean search takes a median of 1.7\,s and a mean of
16.7\,s per design on the 1,405 static-filter designs, and 7.8\,s and
67.6\,s on the remaining 320 designs (Table~\ref{tab:supp_time_gallery}).
Running the trained policy alone takes a median of 6.58\,s on the 982
release designs and 47.53\,s on the 320, averaging 9.99 kernel
evaluations per design on the 982 at $V=100$.
Table~\ref{tab:supp_time_params} lists runtime and learned parameter
counts for each method; external methods ran on GPUs, whereas ours ran
on CPU workers.
The Time column of main-paper Table~1 reports the mean time per part on
the 348 CADBench parts, matching the 348-part columns of
Table~\ref{tab:supp_time_params}. For our mesh rows this measures search
over the selected solid, and for our drawing row it measures prediction
time. CADENA's rollout time on these parts is 87.7\,s, which is not
printed in main-paper Table~1. The end-to-end columns of
Table~\ref{tab:supp_time_params} report total time from input to
returned program on the 748 parts. On the 701 parts with an input mesh,
the mesh route takes a median of 1,299\,s and a mean of 1,455\,s per
part. Because process start-up is not recorded, these figures are lower
bounds. Our mesh route is small in parameters but not fast: end to end, it
takes a mean of 1,455\,s per part on CPU workers, against 10 to 78\,s
on a GPU for CADENA, CAD-Recode, cadrille-RL and Ortho2CAD RL, 177\,s
for StepCAD and 1,860\,s for CADFit
(Table~\ref{tab:supp_time_params}).
Across all 748 parts, the B-rep route averages 331.2\,s per part with a
median of 94.8\,s on CPU workers (Table~\ref{tab:supp_time_params},
end-to-end columns).
Each row of main-paper Table~1 uses the following learned networks. The
mesh and point rows combine the point network (11.74M), the
candidate-stage face policy (2.75M), and the trained B-rep policy
(2.74M), totaling 17.2M parameters. The B-rep row uses only the trained
policy, and the drawing row uses only the correction network (1.67M).
Main-paper Fig.~1 also shows the completion network (11.83M), which
appears only in Carrier-route ablations and diagnostics
(Table~\ref{tab:supp_ablation}, block B1, and Tables~\ref{tab:supp_mesh_detail}
and~\ref{tab:carrier_code}) and is not used by any row of
main-paper Table~1. We describe the candidate-stage face policy, which
Fig.~1 omits, in Sec.~\ref{sec:supp_mesh}.

\begin{table*}[t]
\centering\small
\setlength{\tabcolsep}{5pt}
\begin{tabular}{@{}lrrrrrrrr@{}}
\toprule
Recovered within (seconds) & 1 & 5 & 10 & 30 & 60 & 120 & 300 & 600\\
\midrule
\multicolumn{9}{@{}l}{\emph{All 1,725 test designs}}\\
Gallery agent (released, our runtime) & 0 & 223 & 476 & 760 & 912 & 980 & 995 & 998\\
\ours Boolean search, face extrusions only & 699 & 850 & 872 & 884 & 887 & 887 & 887 & 888\\
\ours Full solver, live run & 612 & 1,236 & 1,329 & 1,422 & 1,464 & 1,477 & 1,487 & 1,490\\
\midrule
\multicolumn{9}{@{}l}{\emph{938 designs with at least two extrusions in the designer's sequence}}\\
Gallery agent (released, our runtime) & 0 & 61 & 148 & 252 & 289 & 310 & 323 & 326\\
\ours Boolean search, face extrusions only & 318 & 439 & 457 & 468 & 471 & 471 & 471 & 471\\
\ours Full solver, live run & 206 & 552 & 615 & 682 & 711 & 720 & 727 & 729\\
\bottomrule
\end{tabular}
\caption{\textbf{Recovery at the same wall time on the Fusion~360
Gallery.} A design counts within $T$ seconds when the method's record
reaches its volume-exact program within $T$ seconds: the agent and
face-extrusion search within $V\leq100$, the full solver at $V=100$;
the agent's programs are confirmed by fresh replay. Failures and runs
that reach a limit count as not recovered. Face-extrusion search is the
first row of Table~\ref{tab:supp_families}. The seconds come from the
records of separate runs, not from one controlled timing run, and the
agent's seconds include its network calls; its fastest volume-exact
design takes 3.0\,s. Each method stops at its own limits (100
environment calls for the agent, wall limits for our searches), so the
counts stop growing there. Table~\ref{tab:supp_time_gallery} counts the
total wall time of each design's run instead.}
\label{tab:supp_time_x}
\end{table*}

\paragraph{Same wall time, face extrusions only.}
Because the released Gallery agent uses only face extrusions,
Table~\ref{tab:supp_time_x} compares it at the same wall times against
Boolean search restricted to face extrusions and against the full
solver. Within 60\,s, face-extrusion search and the agent recover
similar numbers of designs, 887 against 912, with 305 recovered only by
the search and 330 only by the agent (exact McNemar test, $p=.34$).
Within 600\,s, the agent recovers more designs, 998 against 888 (253
against 363, $p=1.1\times10^{-5}$). On the 938 designs whose designer
sequence has at least two extrusions, face-extrusion search leads at
both limits: 471 against 289 within 60\,s (247 against 65,
$p=3.9\times10^{-26}$) and 471 against 326 within 600\,s (221 against
76, $p=1.4\times10^{-17}$). The agent's overall lead therefore comes
from designs with fewer than two extrusions. When face-extrusion search
succeeds, it finishes quickly: it recovers 699 designs within 1\,s and
has a median time of 0.51\,s over recovered designs, compared with
11.1\,s for the agent and 1.44\,s for the full solver. At every time
limit on both sets, the full solver recovers more designs than the
agent.

\begin{table}[t]
\centering\footnotesize
\setlength{\tabcolsep}{3pt}
\begin{tabular}{@{}lrrr@{}}
\toprule
Per design, $V=100$ & Mean & Median & P90\\
\midrule
Modeling operations & 394 & 124 & 1,231\\
\quad of which Booleans & 148 & 46 & 504\\
All kernel calls & 98,626 & 26,533 & 251,767\\
\bottomrule
\end{tabular}
\caption{\textbf{Every kernel operation of the full solver} during a live
run on the 1,725 Fusion~360 Gallery test designs. Modeling operations
consist of Booleans; prism, revolve, and primitive construction; fillet,
chamfer, and shell; defeaturing; and unify and copy. All kernel calls
also count point classifications, classifier setup, and geometry queries.
Only replays of new program prefixes are charged to $V$; the remaining
operations are not charged, though they are counted here.}
\label{tab:supp_kernelops}
\end{table}

Running the full solver live recovers 1,406 of the 1,725 Gallery designs
at $V=20$ and 1,490 at $V=100$, compared with 1,408 and 1,492 composed
from the standalone runs (Table~\ref{tab:supp_brep_full}); the two
designs that differ at $V=100$ hit the wall limit in the live run.
Counting every kernel call, the full solver performs a median of 124
modeling operations per design (46 of them Booleans) at both $V=20$ and
$V=100$, with means of 286 and 394 (Table~\ref{tab:supp_kernelops}).
Because $V$ counts only the program prefixes replayed for checking, most
of the other calls are point classifications that set up the probes.

\subsection{Stand-alone CadQuery programs}
\begin{table}[t]
\centering\footnotesize
\setlength{\tabcolsep}{3pt}
\begin{tabular}{@{}lrr@{}}
\toprule
Programs & $n$ & Script within $10^{-5}$\\
\midrule
Returned volume-exact programs & 591 & 588 (99.5\%)\\
Returned best states (part not solved) & 86 & 61 (70.9\%)\\
All returned programs & 677 & 649 (95.9\%)\\
\midrule
Every volume-exact program found & 1,425 & 1,411 (99.0\%)\\
\midrule
\multicolumn{3}{@{}l}{\emph{Test parts, all returned programs}}\\
Gallery (1,725) & 1,719 & 1,682 (97.8\%)\\
748 parts of main-paper Table~1 & 560 & 510 (91.1\%)\\
CADBench DeepCAD & 2,997 & 2,954 (98.6\%)\\
CADBench Fusion~360 & 2,943 & 2,619 (89.0\%)\\
CADBench ABC Extrude & 2,853 & 2,695 (94.5\%)\\
CADBench ABC All-Ops & 2,882 & 2,258 (78.3\%)\\
\bottomrule
\end{tabular}
\caption{\textbf{Stand-alone CadQuery programs.} A deterministic emitter
writes each B-rep program as a CadQuery script using literal sketch
coordinates, extents, axes, radii, and angles, with no file import. Each
script runs alone in a fresh process, and we compare its solid against
the interpreter's replay of the same program (kernel volume IoU within
$10^{-5}$). Upper block: programs returned by the B-rep export for 678
training-side validation parts (one part has no output). Lower block:
all programs returned on test parts with no limit on $V$ (for the 748
parts, those returned within $V=100$). Here $n$ counts the parts with a
returned program, and any program the emitter does not write counts
against the share.}
\label{tab:supp_cqemit}
\end{table}

B-rep programs are action lists whose operands reference faces of $T$. A
deterministic emitter writes each program as a stand-alone script for
CadQuery~\cite{cadquery}, a Python CAD library built on the Open CASCADE
kernel. The emitter replays the program in our interpreter, records the
arguments of every kernel constructor (prism, half-space, cylinder,
revolve, fillet, chamfer, and thick solid), and outputs each action with
literal sketch curves (lines, three-point arcs, circles, elliptical
arcs, and B-spline edges) in the sketch plane's frame alongside literal
extents, axes, radii, and angles. Finishing operations pick edges and
faces using literal points. We write a script only when running it step
by step beside the interpreter reproduces the interpreter's solid. On
training-side validation parts, 588 of 591 volume-exact programs replay
from the script alone within $10^{-5}$; across all volume-exact
programs found on training-side parts, 1,411 of 1,425 do so
(Table~\ref{tab:supp_cqemit}). On test parts, scripts replay within
$10^{-5}$ for 1,682 of the 1,719 programs returned on the Gallery
(97.8\%) and for 510 of the 560 returned on the 748 parts (91.1\%). On
the CADBench splits, this share ranges from 78.3\% on ABC All-Ops to
98.6\% on DeepCAD.
The median script contains 35 lines and 42 numeric literals. These
scripts carry no sketch constraints; as main-paper Sec.~5 states, a
program that reproduces the target geometry need not recover the
original modeling history or its design constraints. By construction,
section-stack and drawing programs are already written as stand-alone
CadQuery scripts.

\begin{table}[t]
\centering\footnotesize
\setlength{\tabcolsep}{3pt}
\begin{tabular}{@{}lrr@{}}
\toprule
& Interpreter & Scripts\\
\midrule
\multicolumn{3}{@{}l}{\emph{Gallery (1,725), volume-exact to the input}}\\
$V=20$ & 1,408 & 1,391\\
$V=100$ & 1,492 & 1,471\\
\midrule
\multicolumn{3}{@{}l}{\emph{B-rep row of main-paper Table~1 (748 parts)}}\\
IND & .736 & .681\\
Parts with IND $\geq.95$ & 503 & 475\\
\midrule
\multicolumn{3}{@{}l}{\emph{Mesh row of main-paper Table~1 (748 parts)}}\\
IND & .864 & .839\\
Parts with IND $\geq.95$ & 609 & 590\\
\bottomrule
\end{tabular}
\caption{\textbf{Main-paper Tables~1 and~2 scored on the stand-alone
scripts.} Interpreter: the solid of our interpreter's replay, as in the
main paper. Scripts: the solid of the program's stand-alone CadQuery
script, run alone in a fresh process; a program without a script, or
whose script does not replay, counts as a failure. Gallery: designs
recovered by the full solver within $V$ whose exported program's
script is volume-exact to the input. B-rep row: every output of the
row, scored with the row's own scorer. Mesh row: the scripts of the
360 returned search programs, scored with the row's own scorer; the
returned section stacks are scripts already and keep their scores, and
parts without an output score zero.}
\label{tab:supp_script_scores}
\end{table}
Table~\ref{tab:supp_script_scores} scores main-paper Tables~1 and~2
using the solids of the stand-alone scripts. On the Gallery, 1,391
designs (80.6\%) at $V=20$ and 1,471 (85.3\%) at $V=100$ yield a script
whose solid is volume-exact to the input, compared with 1,408 and 1,492
under the interpreter. Among the 21 designs lost at $V=100$, 8 have no
script and 13 produce a script solid that is not volume-exact to the
input. On the 748 parts, we write every output of the B-rep row as a
script, run it alone, and score it with the row's scorer. Mean IND drops
from .736 to .681, and the count of parts with IND of at least .95 falls
from 503 to 475. Out of the 674 parts with a program, 557 produce a
script that builds a valid solid. The drop comes from programs whose
script is missing or fails to replay, mostly partial programs returned
for parts without a volume-exact program. Across the 530 parts whose
script reproduces the interpreter's solid within $10^{-5}$, the two IND
scores agree within $10^{-3}$ on 524. Without a limit on $V$, 417 of
these scripts are volume-exact to the input, whereas main-paper Table~1
counts 403 within $V=100$.
We rescore the mesh row in the same way. Its 337 section-stack programs
are CadQuery scripts by construction and keep their scores, while its 51
parts without an output score zero. Running the same emitter on its 360
search programs produces scripts for 345 parts and none for 15; every
written script replays, and 335 reproduce the interpreter's solid within
$10^{-5}$. Scored on script solids, with every program lacking a script
scored zero, mean IND across the 748 parts falls from .864 to .839, and
the share of parts with IND of at least .95 drops from .814 to .789 (609
to 590 parts). Over the 360 parts with a search program, mean IND drops
from .968 to .915.

\begin{table}[t]
\centering\footnotesize
\setlength{\tabcolsep}{3pt}
\begin{tabular}{@{}lrr@{}}
\toprule
& Gallery & 748 parts\\
\midrule
\multicolumn{3}{@{}l}{\emph{Scripts volume-exact to the input}}\\
Programs & 1,552 & 393\\
Actions per program, median / mean & 2 / 2.30 & --\\
Human extrusions, median / mean & 2 / 2.02 & --\\
Shorter than the human sequence & 21.5\% & --\\
Equal & 54.5\% & --\\
Longer & 24.0\% & --\\
\midrule
\multicolumn{3}{@{}l}{\emph{Share of programs using each family, \%}}\\
X / XD & 69.6 / 10.1 & 88.3 / 9.9\\
H / R & 21.6 / 32.2 & 8.9 / 20.9\\
F / C & 0.6 / 1.8 & 6.6 / 1.0\\
S / P & 0 / 0.1 & 0 / 0.3\\
\midrule
\multicolumn{3}{@{}l}{\emph{First dimension $+10\%$, script run again}}\\
Scripts written & 1,695 & 529\\
Scripts with a dimension edited & 1,440 & 473\\
Edited script returns a solid & 1,393 & 451\\
Valid solid & 1,384 & 447\\
Valid solid with a changed volume & 1,270 & 403\\
\bottomrule
\end{tabular}
\caption{\textbf{Program size and editing.} Top: solver actions per
returned program with a script volume-exact to the input (Gallery: no
limit on $V$; 748 parts: programs returned within $V=100$), compared
with the extrusions in the Gallery's human design sequences. Our
actions also include holes, fillets, and revolves, which the extrusion
count omits. Middle: share of these programs that use each action
family (Table~\ref{tab:supp_actions}); each program is counted once
per family it uses. Bottom: the first dimension literal in each script
(an extrusion distance, hole radius, revolve angle, or fillet radius)
is increased by 10\%, and the script is run again. Validity is checked
by the kernel's shape check, and we make no claim about the intended
effect of the edit. --: not computed; the 748 parts have no human
design sequences.}
\label{tab:supp_edit}
\end{table}

Measured in solver actions, the 1,552 Gallery programs whose scripts are
volume-exact to the input have lengths close to the designers' extrusion
sequences: both have a median of 2, with our program shorter on 21.5\%
of designs, equal on 54.5\%, and longer on 24.0\%
(Table~\ref{tab:supp_edit}). Measured in CADBench's CadQuery operations,
our scripts are long because every sketch curve is written as a literal.
On the 276 CADBench parts among the 748 that have a script, the median
is 68 operations (72 for volume-exact scripts), compared with 16 for
cadrille-RL, 29 for CAD-Recode, 42 for Ortho2CAD RL, 48 for StepCAD, and
77 for CADFit on the same parts. When we raise the first dimension of
each script by 10\%, 1,270 of the 1,440 edited Gallery scripts and 403
of the 473 on the 748 parts rebuild into a valid solid with a changed
volume.

\begin{table}[t]
\centering\footnotesize
\setlength{\tabcolsep}{3pt}
\begin{tabular}{@{}lrrr@{}}
\toprule
& \multicolumn{2}{c}{Ours} & \\
\cmidrule(lr){2-3}
Share of 200 programs, \% & Search & Stacks & CAD-Recode\\
\midrule
A dimension is edited & 83.5 & 89.5 & 94.5\\
Rebuilt: returns a solid & 81.5 & 89.0 & 91.0\\
Valid & 81.5 & 82.5 & 71.0\\
Intended change & 73.5 & 88.5 & 88.0\\
Other faces preserved & 75.0 & 84.0 & 90.5\\
All three & 67.5 & 77.5 & 68.5\\
\bottomrule
\end{tabular}
\caption{\textbf{Named-edit test.} In each program, the first hole
radius, or else the first extrusion distance, is raised by 10\% and the
script is run again. Valid: the kernel's shape check passes. Intended:
the named feature changes by exactly the edit (the hole's radius, or
the extent of the extruded body). Preserved: every face not produced by
the edited operation keeps its surface. All three: valid, intended and
preserved. Programs without such a dimension count as failures.
Search: our B-rep search programs on the Gallery and the 748 parts,
with a script that reproduces the program's solid. Stacks: our mesh
route's section stacks on the 748 parts. CAD-Recode
v1.5: its programs on the 748 parts. Each group holds the first 200
programs of a seeded draw whose unedited script runs.}
\label{tab:supp_named_edit}
\end{table}
The test in Table~\ref{tab:supp_edit} checks validity only, whereas
Table~\ref{tab:supp_named_edit} also checks that each edit produces its
named effect across 200 programs per group. All three checks pass on
67.5\% of our B-rep search programs, 77.5\% of our section stacks, and
68.5\% of CAD-Recode v1.5 programs. Under this test, no group is clearly
more editable: section stacks pass most often, while our search programs
and CAD-Recode perform similarly. By edit type, all three checks pass
for 63 of 85 hole radii and 72 of 82 extrusion distances in our search
programs, 70 of 85 and 85 of 94 in section stacks, and 60 of 90 and 77
of 99 in CAD-Recode programs. Our search programs fail most often on the
intended change of a hole radius, whereas CAD-Recode fails most often on
validity, with 40 of its 182 rebuilt solids invalid.

\begin{table}[t]
\centering\footnotesize
\setlength{\tabcolsep}{3pt}
\begin{tabular}{@{}lrrr@{}}
\toprule
& \multicolumn{2}{c}{Ours} & \\
\cmidrule(lr){2-3}
Parts, of 150 & Search & Stacks & CAD-Recode\\
\midrule
\multicolumn{4}{@{}l}{\emph{Drawn from parts with a program from all three}}\\
Pass & \textbf{57} & 49 & 37\\
No matching element & 59 & 71 & 94\\
No runnable script & 17 & 18 & 11\\
Fails a check & 17 & 12 & 8\\
\midrule
\multicolumn{4}{@{}l}{\emph{Drawn from all 748 parts}}\\
Pass & 44 & \textbf{52} & 32\\
No matching element & 44 & 63 & 94\\
No runnable script & 38 & 23 & 23\\
Fails a check & 24 & 12 & 1\\
\bottomrule
\end{tabular}
\caption{\textbf{The same edit on the same parts.} The edit is defined
from the reference part: its largest through-hole radius, or else its
extent along the main extrusion axis, is raised by 10\%. A program must
contain the matching element (a hole of that radius on that axis, or an
extrusion spanning that extent); the edited script must then rebuild a
valid solid with the intended change and every other face preserved.
Search: our B-rep search programs, from the B-rep. Stacks: the
section-stack candidate our mesh route builds for the part, from the
mesh. CAD-Recode v1.5: its program from 256 points. In the lower block, a part without a program counts as having no
runnable script.}
\label{tab:supp_paired_edit}
\end{table}

\paragraph{The same edit on the same parts.}
In Table~\ref{tab:supp_named_edit}, the groups contain different parts, and
each program's own first hole radius or extrusion distance is edited. Table~\ref{tab:supp_paired_edit}
instead defines one edit from the reference part and applies it to every method
on the same parts. Across 150 parts drawn from those where all three methods
return a program, our B-rep search programs pass on 57 (38\%), section stacks
pass on 49 (33\%), and CAD-Recode v1.5 passes on 37 (25\%). Our search programs
pass where CAD-Recode fails on 38 parts, while the reverse happens on 18 parts
(exact McNemar test, $p=.010$). Against section stacks, the split is 20 to 12
($p=.22$). Across 150 parts drawn from all 748, the pass counts are 44 (29\%),
52 (35\%), and 32 (21\%), respectively. On this sample, section stacks beat
CAD-Recode 34 to 14 ($p=.006$), our search programs beat CAD-Recode 26 to 14
($p=.081$), and search programs and section stacks split 10 to 18 ($p=.18$).
For every method, the most common failure is a program with no element matching
the reference feature, such as no single extrusion spanning the part's height.
On the shared parts, the hole edit passes on 34 of 68 parts for our search
programs and 35 of 68 for section stacks. The height edit passes on 23 of 82
parts for our search programs and 14 of 82 for section stacks.

\subsection{Ablations}
\begin{table}[t]
\centering\scriptsize
\setlength{\tabcolsep}{3pt}
\begin{tabular}{@{}lrrrr@{}}
\toprule
\multicolumn{5}{@{}l}{\textbf{A. What expert iteration adds} (ABC-EVAL, volume-exact at $V=24$)}\\
& 533 files & 400 files & & \\
\midrule
Imitation policy & 155 & 104 & & \\
\ours Trained policy & \textbf{174} & \textbf{120} & & \\
Gains / losses & 22 / 3 & 18 / 2 & & \\
Paired exact test, $p$ & .00016 & .0004 & & \\
\midrule
\multicolumn{5}{@{}l}{\textbf{B1. Carrier route, one part at a time} (ABC-EVAL, 533)}\\
& IND & \#$\geq.95$ & & \\
\midrule
Cells from fitted surfaces only & .532 & 194 & & \\
+ analytic extrude and revolve proposals & .573 & 225 & & \\
+ learned completion & .582 & 236 & & \\
\ours + guarded face merging (Carrier route) & \textbf{.635} & \textbf{249} & & \\
\midrule
\multicolumn{5}{@{}l}{\textbf{B2. Selecting among candidates} (ABC-EVAL, 533)}\\
& IND & OSC & FeatR95 & Verified\\
\midrule
Carrier route, no merging, no selection & .582 & .619 & .450 & 67\\
Priority selector & .729 & .658 & .498 & 77\\
Restricted pool (no section stacks) & .693 & .508 & .317 & 61\\
Section stacks only & .781 & .737 & .655 & 162\\
\ours Full pool & \textbf{.834} & \textbf{.789} & \textbf{.693} & \textbf{167}\\
\emph{Input mesh returned (reference)} & \emph{.856} & \emph{.950} & \emph{1.000} & \emph{442}\\
\midrule
\multicolumn{5}{@{}l}{\textbf{C. What code length explains} (Gallery 982)}\\
& \multicolumn{2}{c}{Spearman with calls} & \multicolumn{2}{c}{Median bits}\\
& code & depth & imit. & trained\\
\midrule
Programs needing $>1$ call ($n=231$) & \textbf{.868} & .418 & & \\
All recoveries & .677 & \textbf{.839} & & \\
Programs found by the trained policy & & & 1.76 & .09\\
Programs found by Boolean search & & & 5.05 & 7.58\\
\midrule
\multicolumn{5}{@{}l}{\textbf{D. Policy search settings} (ABC-EVAL 533, volume-exact at $V=24$)}\\
& on & off & $p$ & \\
\midrule
Pooling, trained policy & 174 & 166 & .008 & \\
\bottomrule
\end{tabular}
\caption{\textbf{Ablations grouped by component.}
A: paired exact tests on all 533 files (first STEP root) and on the 400
single-solid files. B1: Carrier-route components added one at a time.
B2: candidate-solid selection rules. Verified counts outputs that pass
the boundary check (out of 476--481 eligible parts), whether exact or
not. The Full pool selects by fit to the input mesh. The input mesh is a
tessellation of the reference; returning it serves as an upper reference
that produces no solid or program. C: rank correlation between policy
calls and either the trained policy's conditional code length or program
depth, together with the median conditional code length of the same
programs under both policies. D: policy search with and without pooling
sibling actions that build the same tool with the same operation (8 and
0 discordant parts). Shading marks our default configurations. Blocks B1 and B2 come from an
earlier checkpoint of the mesh route (as in Table~\ref{tab:supp_mesh_detail});
the learned completion in B1 is not used by any row of main-paper Table~1.}
\label{tab:supp_ablation}
\end{table}

Table~\ref{tab:supp_ablation} groups the ablations by component. Expert
iteration improves policy-only recovery at small budgets (A), though at
$V=100$ its gain on the 533 files narrows to 181 against 173 (12 gains
and four losses, $p=.077$). Each Carrier-route component raises mean
IND (B1), though the gain from learned completion ($+.0085$, 95\% CI
$[-.0095,+.0263]$) is within noise. Among the candidate rules, selecting by input fit
gives the highest overlap and boundary scores (B2).
Block C links code length to search effort on the 982 release designs
and supplies the medians in main-paper Sec.~4.4 (1.76, 0.09, and 7.58
bits). Here, 7.58 bits is the median under the trained policy across all
programs that Boolean search finds on these 982 designs, including
designs that the trained policy also recovers rather than only those
found by Boolean search alone. Under the imitation policy, the same
programs take 5.05 bits.

\paragraph{Expert iteration on training-side parts.}
On 512 Zero-to-CAD validation parts, the imitation policy solves 183
parts at 24 calls and 234 at 100 calls. The trained policy solves 258
parts at 100 calls, and its 24-call count already exceeds the imitation
policy's 100-call count. On 278 ABC training-validation parts, the
trained policy solves 171 against 154 at 24 calls (22 gains and five
losses) and 180 against 171 at 100 calls. Most newly solved parts were
near misses before training. Across the 512 Zero-to-CAD parts, recovery
drops from 71\% for 7--16 faces to 38\% above 30 faces.

\begin{table}[t]
\centering\footnotesize
\setlength{\tabcolsep}{3pt}
\begin{tabular}{@{}lrrr@{}}
\toprule
Volume-exact within & 8 calls & 24 calls & 100 calls\\
\midrule
\multicolumn{4}{@{}l}{\textbf{A. Thirty more rounds} (512 parts)}\\
\ours Trained policy (start) & 214 & 254 & 282\\
Plain recipe & 212 & 260 & 282\\
Partial credit & 210 & 252 & 277\\
Step contrast & 204 & 250 & 279\\
Partial credit and step contrast & 207 & 253 & 280\\
Corrected coverage input & 218 & 260 & 282\\
\midrule
\multicolumn{4}{@{}l}{\textbf{B. Anti-collapse recipe} (503 parts)}\\
Imitation policy & -- & 180 & 231\\
\ours Trained policy & -- & 234 & 254\\
Anti-collapse recipe & -- & 212 & 243\\
\midrule
\multicolumn{4}{@{}l}{\textbf{C. Policy size} (imitation policies, 512 parts)}\\
2.7M parameters & -- & 183 & 234\\
39M parameters & -- & 187 & --\\
\bottomrule
\end{tabular}
\caption{\textbf{Variants of expert iteration (training-side).}
Volume-exact solves on the 512 Zero-to-CAD training-validation parts.
A: each
arm starts from the trained policy and trains 30 more rounds on the
same targets, with the plain recipe; with partial credit, which also
imitates the best partial program of a failed search; with a step
contrast loss on the evaluated children of expanded states; with both;
or with a corrected coverage input to the policy. B: a recipe that
learns from search solves except those found at the first call, fills
half of each batch with imitation data and imitates the synthetic
program of a failed target; parts with a record for all three
policies. C: imitation policies of two sizes. Each block is a separate
evaluation; parts without a valid record count as unsolved. The search
of block A skips repeated states and finishing actions that select
nothing; the text's count for the trained policy (258 at 100 calls) comes
from a search without these two checks. --: not
reported.}
\label{tab:supp_exit_variants}
\end{table}

\paragraph{Variants of expert iteration (training-side).}
Table~\ref{tab:supp_exit_variants} compares alternative training recipes
on the same 512 training-validation parts. In block A, no arm solves
more parts at 100 calls than the starting trained policy. At 24 calls,
neither the plain recipe ($p=.23$) nor the corrected coverage input
($p=.12$) differs significantly from the starting policy in a paired
test. Partial credit, the step contrast loss, and both together
solve no more parts than the plain recipe at any budget. The step
contrast loss does alter how search spends its calls: in 100-call
searches, the share of calls that lose a built target face or change
nothing drops from 37.7\% to 29.7\%. The training searches provide a
larger comparison because every arm searches the same fresh targets in
each round, about 5,000 per arm in total. Against the plain recipe,
partial credit solves 2,732 against 2,728 of 5,015 targets, the step
contrast loss solves 2,748 against 2,738 of 5,120, and the corrected
input solves 2,753 against 2,738 of 5,127. The anti-collapse recipe
solves fewer parts than the trained policy, gaining 13 and losing 35 at
24 calls ($p=.0021$) and gaining 13 while losing 24 at 100 calls
($p=.099$). A 39M imitation policy solves 187 parts at 24 calls,
compared with 183 for the 2.7M policy. None of these variants clearly improves
on the plain recipe.

\paragraph{Policy size (training-side).}
Table~\ref{tab:supp_policy_size} evaluates the B-rep policy at four
sizes, from 2.74M to 125M parameters, trained on the same demonstrations
and searched with the same budget on 293 training-side validation parts.
Although validation loss falls with size from about 1.10 to about 1.03,
exact recoveries at 100 evaluations stay between 122 and 138, whereas
three seeds of the smallest policy alone span 122 to 135. At 24
evaluations, the largest policy is ahead (115 against 105 to 110). On
300 additional ABC training parts, all four sizes recover 196 to 203
parts. A 1.26M policy also matches the 2.74M one, and the 39M policy in
Table~\ref{tab:supp_exit_variants} (block C) gains 4 parts. We therefore
use the 2.74M policy. We did not compare sizes for the point,
completion, or drawing networks.

\begin{table}[t]
\centering\footnotesize
\setlength{\tabcolsep}{3pt}
\begin{tabular}{@{}lrrr@{}}
\toprule
Parameters & Val.\ loss & Exact, $V{=}24$ & Exact, $V{=}100$\\
\midrule
2.74M (3 seeds) & 1.100 / 1.123 / 1.092 & 110 / 105 / 108 & 135 / 126 / 122\\
12.4M & 1.049 & 106 & 131\\
39.1M (2 seeds) & 1.056 / 1.045 & 106 / 109 & 133 / 134\\
125M (2 seeds) & 1.029 / 1.041 & 115 / 115 & 138 / 132\\
\bottomrule
\end{tabular}
\caption{\textbf{B-rep policy size (training-side).} The same design and
demonstrations at four sizes; volume-exact recoveries out of 293
training-side validation parts under the same search. Three seeds of
the smallest policy span 122 to 135 at $V{=}100$.}
\label{tab:supp_policy_size}
\end{table}

\begin{table*}[t]
\centering\small
\setlength{\tabcolsep}{5pt}
\begin{tabular}{@{}lrrrrrrr@{}}
\toprule
& \multicolumn{2}{c}{Gallery (1,725)}
& \multicolumn{3}{c}{748 parts of main-paper Table~1}
& \multicolumn{2}{c}{CADBench (3,000 each), $V=100$}\\
\cmidrule(lr){2-3}\cmidrule(lr){4-6}\cmidrule(l){7-8}
Tool families & $V=20$ & $V=100$ & $V=20$ & $V=100$ & No limit & DeepCAD & Fusion~360\\
\midrule
Face extrusions (X) & 865 & 890 & 315 & 323 & 326 & 2,850 & 1,219\\
+ revolves (R) & 1,247 & 1,273 & 375 & 388 & 393 & 2,890 & 1,711\\
+ holes (H) & 1,260 & 1,288 & \textbf{379} & 394 & 399 & 2,914 & 1,731\\
+ defeaturing, hole finishing & 1,346 & 1,393 & 348 & 363 & 388 & 2,910 & 1,789\\
\ours + XD bodies; F, C, S finishing &\textbf{1,382} & \textbf{1,431} & 365 & \textbf{400} & \textbf{424} & \textbf{2,924} & \textbf{1,896}\\
\bottomrule
\end{tabular}
\caption{\textbf{Tool families added one at a time.} All runs use
Boolean search from the B-rep without the policy. Entries report the
number of volume-exact programs found within $V$ kernel evaluations,
using the accounting of Table~\ref{tab:supp_brep_full}; ``No limit''
counts volume-exact programs found within the same wall limit with no
cap on $V$. Each row adds its tool families to those above it. Face
extrusions and revolves are the tool families from prior
work~\cite{willis2021fusion360,xu2021zonegraph,yin2025rlcad}; the rows
below add the families introduced in this work. The last row adds
extrusions of the defeatured body (XD) and fillet, chamfer, and shell
finishing (F, C, S); this gives the full Boolean search. The CADBench
columns report the DeepCAD and Fusion~360 splits from the STEP files.}
\label{tab:supp_families}
\end{table*}

When we add tool families to Boolean search one at a time without the
policy, revolves give the largest single gain on every set
(Table~\ref{tab:supp_families}). At $V=100$, the full tool set recovers
1,431 Gallery designs compared with 890 using face extrusions alone, and
1,896 of the 3,000 CADBench Fusion~360 designs compared with 1,219. On
CADBench DeepCAD, face extrusions alone already recover 2,850 of 3,000.
On the 748 parts, defeaturing with only hole finishing lowers recovery
from 394 to 363, while the full tool set recovers 400 at $V=100$ and 424
without a limit on $V$.
Face extrusions and revolves are the tool families from prior work
(Sec.~\ref{sec:supp_brep}, Actions). Together, the families added here
(holes, XD bodies, and fillet, chamfer, and shell finishing) raise
recovery at $V=100$ from 1,273 to 1,431 Gallery designs, from 388 to 400
on the 748 parts, and from 1,711 to 1,896 CADBench Fusion~360 designs.
Adding the trained policy raises these counts to 1,492, 403, and 1,968
(Table~\ref{tab:supp_brep_full}, main-paper Table~1, and
Table~\ref{tab:supp_cadbench}).
Shell finishing appears in none of the programs whose scripts are
volume-exact to the input, neither in the 1,552 on the Gallery nor in
the 393 on the 748 parts. Fillets appear in 0.6\% and 6.6\% of them, and
chamfers in 1.8\% and 1.0\% (Table~\ref{tab:supp_edit}).

\begin{table}[t]
\centering\footnotesize
\setlength{\tabcolsep}{3pt}
\begin{tabular}{@{}lrr@{}}
\toprule
& Probe bits & Kernel\\
\midrule
States scored per second, median & $\approx$990,000 & 12.4\\
\midrule
\multicolumn{3}{@{}l}{\emph{Boolean search, 200 test designs, same wall limits}}\\
Volume-exact, $V=20$ & \textbf{120} & 46\\
Volume-exact, $V=100$ & \textbf{128} & 71\\
Volume-exact, no limit on $V$ & \textbf{135} & 105\\
Seconds per design, mean / median & \textbf{127} / \textbf{4.6} & 336 / 147\\
\bottomrule
\end{tabular}
\caption{\textbf{Probe screening against kernel scoring.} Top:
candidate states scored per second on 50 training-side Gallery parts,
either with probe bits or by building the state in the kernel and
computing its volume IoU with the target. Bottom: Boolean search on 200
test designs (100 from the 1,725 Gallery designs and 100 from the 748
parts in main-paper Table~1). Both methods use the same tools, beam
width (512), depth, stages, wall limits, and accounting; only
candidate-state scoring differs. Every state built in the kernel is a
kernel evaluation and counts toward $V$. We did not try a narrower
kernel beam.}
\label{tab:supp_probes}
\end{table}

Under the same wall limits, scoring candidate states on probe points
instead of building them in the kernel raises recovery on 200 test
designs from 105 to 135 volume-exact programs without a limit on $V$,
and from 71 to 128 at $V=100$ (Table~\ref{tab:supp_probes}). The probe
beam evaluates about $10^5$ times more states per second; on most
designs, the kernel-scored beam finishes only its first one or two
levels before hitting the wall limit.

\section{CADBench}
\label{sec:supp_cadbench}

\begin{table*}[t]
\centering\small
\setlength{\tabcolsep}{4pt}
\begin{tabular}{@{}lcccc@{}}
\toprule
Row & DeepCAD & Fusion~360 & ABC Extrude & ABC All-Ops\\
\midrule
\multicolumn{5}{@{}l}{\emph{Mesh input, published}~\cite{doris2026cadbench}: median aligned IoU}\\
CADFit~\cite{nehme2026cadfit}$^\P$ & .986/.966/.919 & .983/.955/.868 & .973/.835/.835 & .934/.789/.789\\
CADEvolve~\cite{elistratov2026cadevolve} & .970/.924/.824 & .917/.827/.605 & .928/.548/.621 & .855/.541/.480\\
CAD-Recode~\cite{rukhovich2025cadrecode} & .963/.908/.804 & .917/.816/.546 & .917/.424/.540 & .837/.485/.415\\
cadrille~\cite{kolodiazhnyi2026cadrille} & .961/.904/.808 & .911/.813/.584 & .911/.485/.554 & .840/.514/.430\\
\midrule
\multicolumn{5}{@{}l}{\emph{Mesh input, ours: selected solid}}\\
\ours Median aligned IoU & -- & -- & .996/.983/.988 & .993/.972/.964\\
\ours SIoU & -- & -- & .865/.901/.879 & .826/.859/.813\\
\ours VSR & -- & -- & .993/.992/.990 & 1.000/.999/.996\\
\ours IoU, training overlap excluded & -- & -- & .996/.979/.988 & .993/.969/.961\\
\midrule
\multicolumn{5}{@{}l}{\emph{STEP input, ours: returned program}}\\
\ours Median aligned IoU & 1.000/1.000/1.000 & .999/.998/.989 & 1.000/1.000/.999 & .997/.970/.876\\
\ours SIoU & .845/.862/.820 & .881/.877/.817 & .867/.905/.831 & .819/.779/.628\\
\ours CD & .026/.025/.027 & .024/.025/.027 & .025/.023/.025 & .027/.029/.033\\
\ours VSR & .998/1.000/.999 & .954/.995/.984 & .999/.998/.852 & .997/.984/.829\\
\ours Volume-exact programs (of 3,000) & 2,930 & 1,968 & 2,499 & 1,391\\
\bottomrule
\end{tabular}
\caption{\textbf{CADBench, all four splits.} Cells list low/medium/high
face-count tiers, with 1,000 designs per tier. IoU and SIoU are
CADBench's aligned volume IoU and surface IoU, given as tier medians
with failures scored as zero. CD is the median Chamfer distance over
valid outputs, and VSR is the valid-solid rate. Our selected-solid rows
cover all 3,000 designs in each ABC split. We did not run our mesh route
on the CADBench DeepCAD and Fusion~360 splits;
Table~\ref{tab:supp_standard} gives results on the full DeepCAD and
Fusion~360 test meshes instead. The overlap-excluded mesh row uses 1,153
ABC Extrude designs and 474 of the first 1,000 ABC All-Ops designs.
$^\P$CADBench v2 reports identical medium- and high-tier values for
CADFit on both ABC splits; its first version gives .982/.896/.876 on ABC
Extrude and .950/.859/.832 on ABC All-Ops.}
\label{tab:supp_cadbench}
\end{table*}

\paragraph{Protocol.}
We use the four splits of 3,000 designs in
CADBench~\cite{doris2026cadbench} (DeepCAD, Fusion~360, ABC Extrude,
and ABC All-Ops), each divided into three face-count tiers. The
published methods take meshes as input. Our B-rep row takes the
reference STEP file, so it measures program recovery given a known
boundary and answers a different question. To score our exported
solids, we tessellate them and run CADBench's scorer, which reports
tier medians of aligned volume IoU and surface IoU (SIoU) with failures
scored as zero. We compute volume-exact counts using our kernel IoU at
$10^{-5}$ under the shared budget $V=100$.
Table~\ref{tab:supp_cadbench} extends Table~3 of the main paper by
adding the DeepCAD and Fusion~360 splits, SIoU, Chamfer distance, and
the valid-solid rate. On ABC Extrude and ABC All-Ops, the selected mesh
solids are mostly section stacks with many faces, so their IoU reflects
recoverable geometry rather than compact programs. Table~3 of the main
paper reports the returned programs on a 348-design sample.

\paragraph{CADFit cells.}
The published rows in Table~3 of the main paper and in
Table~\ref{tab:supp_cadbench} come from CADBench v2. Version~2 lists
identical medium and high cells for CADFit on both ABC splits
(.835/.835 and .789/.789). Its first version reports .982/.896/.876
(ABC Extrude) and .950/.859/.832 (ABC All-Ops), so the low cells differ
as well (.973 and .934 in version~2). Our mesh rows exceed both
versions of its table.

\paragraph{Training-data overlap.}
Many CADBench ABC designs share an ABC identifier with our training
data. Almost all of this overlap comes from a surface dataset used only
by networks not included in any row of Table~1 in the main paper. We
check overlap by ABC identifier, so near-duplicate shapes under other
identifiers are not detected.
The completion network's training parts include 83 ABC Extrude and 126
ABC All-Ops designs, with two hard designs in each split. We train the
policy only on Zero-to-CAD targets.
The B-rep search uses no surface model. Excluding every overlapping
design leaves 1,153 ABC Extrude designs for the mesh route, with tier
medians of .996/.979/.988 (Table~\ref{tab:supp_cadbench}). These
remaining designs are larger, with a median STEP file size of 1.1
versus .77\,MB on the hard tier. On 474 of the first 1,000 ABC All-Ops
designs, the medians are .993/.969/.961. The first mesh row in Table~3
of the main paper (.996/.983/.988 and .993/.972/.964) includes the
overlapping designs.

\section{Meshes and points}
\label{sec:supp_mesh}

\begin{table}[t]
\centering\small
\setlength{\tabcolsep}{3pt}
\begin{tabular}{@{}lrrrr@{}}
\toprule
Clean mesh & $n$ & Restricted & Full & Stack picks\\
\midrule
\ours DeepCAD & 8,046 & .958 & \textbf{.995} & 3,830\\
\ours Fusion~360 & 1,725 & .917 & \textbf{.992} & 1,252\\
\ours \quad design-disjoint & 787 & .906 & .988 & 546\\
\midrule
256 unoriented points & $n$ & Restricted & Full & Carrier\\
\midrule
\ours DeepCAD & 500 & .776 & .785 & .030\\
\ours Fusion~360 & 500 & .704 & .733 & .023\\
\ours \quad design-disjoint & 216 & .690 & .721 & .011\\
\midrule
8,192 oriented points & $n$ & Restricted & Full & Carrier\\
\midrule
\ours DeepCAD & 500 & .735 & .723 & .496\\
\ours Fusion~360 & 500 & .681 & .696 & .507\\
\midrule
\multicolumn{5}{@{}l}{\emph{Published, own protocol}~\cite{kabisov2026cadena}}\\
CADENA-RL, mesh & & & \multicolumn{2}{r}{.961 / .941}\\
CADFit~\cite{nehme2026cadfit}, mesh & & & \multicolumn{2}{r}{.914 / .882}\\
CAD-Recode~\cite{rukhovich2025cadrecode}, 256 pts & & & \multicolumn{2}{r}{.914 / .872}\\
\bottomrule
\end{tabular}
\caption{\textbf{Solids from the standard DeepCAD and Fusion~360 test
inputs.} Entries report mean IND, with failures scored as zero. Both
pools select by fit to the input mesh, which is a tessellation of the
reference on clean meshes. The Restricted pool excludes section stacks
and patch candidates, and stack picks count the section stacks selected
in the Full pool. Point rows use 500 designs per test set: 256 points
are meshed with an alpha shape, and 8,192 oriented points with Poisson
reconstruction. The Carrier route rarely produces a valid solid from
256 points because many faces receive no samples. Published numbers
(DeepCAD / Fusion~360) are not directly comparable; they report mean IoU
in a fixed normalized frame over valid predictions only (CAD-Recode has
7.9\% and 13.6\% invalid outputs).}
\label{tab:supp_standard}
\end{table}

\begin{table}[t]
\centering\small
\setlength{\tabcolsep}{3pt}
\begin{tabular}{@{}lrrr@{}}
\toprule
Points to program, IND & 748 & 400 & 348\\
\midrule
CAD-Recode v1.5, 256 pts & .482 & .458 & .509\\
cadrille-RL, 256 pts & \textbf{.641} & \textbf{.589} & \textbf{.701}\\
\ours Ours, 256 pts, search only & .091 & .093 & .088\\
\ours Ours, 256 pts, + fallback & .556 & .504 & .615\\
\ours Ours, 8,192 oriented pts & .609 & -- & --\\
\bottomrule
\end{tabular}
\caption{\textbf{Programs from points on the 748 single-solid ABC
parts.} Columns cover all parts, the 400 single-solid files from the
533-file ABC-EVAL set, and the 348 CADBench parts, with failures scored
as zero. The 256-point rows follow the protocol of the point baselines.
CAD-Recode and cadrille decode greedily, producing one program per
part. Adding the section-stack fallback raises IND by .465 over search
alone, mainly on parts where search returns no program. The 8,192-point
row comes from the main paper; 260 of its programs reproduce the
reconstructed solid exactly. --: not split by set.}
\label{tab:supp_points748}
\end{table}

\paragraph{Standard test meshes.}
Across all 8,046 DeepCAD and 1,725 Fusion~360 test meshes, the Full pool
reaches mean IND .995 and .992, while the Restricted pool reaches .958
and .917 (Table~\ref{tab:supp_standard}). Both pools return finite IoU
on 8,010 DeepCAD parts (36 references are non-manifold) and 1,722
Fusion~360 parts. Both have a median output face count of 8 on DeepCAD
and 10 on Fusion~360. Mean face counts for the Full and Restricted pools
are 14.1 and 21.5 on DeepCAD, and 33.2 and 44.4 on Fusion~360. If we
score every output with an out-of-range overlap flag as zero, the Full
and Restricted pools reach .995 and .958 on DeepCAD, and .988 and .913
on Fusion~360. The mesh and point routes use no Fusion~360 training
data. Of these 1,725 test files, 938 come from designs that have other
bodies in the drawing network's training data, though no file is shared;
Table~\ref{tab:supp_standard} reports the remaining 787 separately.

\paragraph{Sparse and dense points.}
With 256 unoriented points, 41\% of faces receive no sample and only
14\% of parts have every face observed on training-side data, so a
method must infer the geometry rather than recover it. In this setting,
the Full pool reaches .785 and .733 on 500 DeepCAD and Fusion~360
designs each, the Carrier route rarely yields a valid solid, and
point-native code predictors score higher
(Table~\ref{tab:supp_standard}). On the 748 ABC parts, programs from
search alone reach .091 IND from 256 points. Adding the section-stack
fallback raises this to .556, below cadrille-RL (.641;
Table~\ref{tab:supp_points748}). Dense input of 8,192 oriented points
produces the main paper's points row: .609 IND on the 748 parts, with
260 programs reproducing their reconstructed solid exactly
($\dagger$ in main-paper Table~1). In the 500-design runs, thirteen
8,192-point parts hit the candidate stage's memory limit and two
DeepCAD references are not manifold; all count as zero.

\paragraph{Points route on the 748 parts.}
On the 748 parts, our point route trails cadrille-RL
in mean IND, .609 against .641, with failures scored as zero. Our
route returns no valid output on 155 parts,
compared with 59 for cadrille-RL. Over parts with a valid output, mean
IND is .768 for our route and .696 for cadrille-RL. When our route
returns a program, it stays closer to the reference: IND reaches .95 on
a .250 share of the parts against .134, and OSC is .320 against .208.
From 256 points, our route reaches .556 against .641
(Table~\ref{tab:supp_points748}). The point route uses no network trained on pairs of point clouds and
CAD programs.

\paragraph{Mesh programs.}
The mesh route considers two candidate programs: the program that B-rep
search finds on the selected solid, and a section-stack program. The
section-stack candidate is either the selected stack's own construction
or, when the Full pool picks a learned solid, the construction of the
best-fitting stack candidate. We replay each candidate and measure its
fit as the mean of the normalized and fixed-frame IoU between the
replayed solid and the input mesh alone. Among candidates whose fit
falls within 0.001 of the best score, the route picks the shortest
program, counting search actions or the extrusions, revolves, and boxes
of a stack. Ties go to the search program. We set this tolerance on 160
training-validation parts.
On the 748 parts, the route returns a section-stack program for 337
parts, a search program for 360, and no program for 51. Search programs
have a median of 1 action and 34 output faces, whereas section-stack
programs have a median of 10 features and 224 faces
(Table~\ref{tab:supp_compact}). Adding the fallback raises IND from
.648 to .864, mainly on parts where search returns no program, while
volume-exact programs rise only from 99 to 102. On parts where the
route returns a search program, mean IND is .968, OSC is .932, and 67\%
of 353 eligible outputs pass the boundary check. Returned section
stacks reach .884 IND and .701 OSC, with 11\% of 325 eligible outputs
passing.
Of the 697 returned programs, 539 replay the route's own reconstructed
solid exactly (the $\dagger$ count in main-paper Table~1), and 102 are
volume-exact against the reference.
We replay each section-stack program with stock CadQuery in an isolated
process and check it against the route's solid.

\paragraph{Section-stack parameters.}
We center the input mesh on its bounding box and express all lengths in
units of the largest bounding-box side. A stack slices the mesh at its
cap levels (the heights of faces perpendicular to the stacking
direction, grouped within $2\cdot10^{-4}$) and extrudes each slab from
its mid-height section. While the largest estimated slab error exceeds
$4\cdot10^{-4}$ of the mesh volume, we split that slab in two, up to 10
slabs. Each section profile is written as a circle, as lines and
circular arcs, or as a polygon. We fit arcs after dropping collinear
vertices, using a radial tolerance of $2\cdot10^{-6}$. Otherwise, we
simplify the polygon with Clipper at tolerance $4\cdot10^{-4}$, raising
the tolerance until at most 240 vertices remain.

\begin{table}[t]
\centering\footnotesize
\setlength{\tabcolsep}{2.5pt}
\begin{tabular}{@{}lrrrrr@{}}
\toprule
Selection setting & IND & $\geq.95$ & Stacks & Search & Features\\
\midrule
\ours Main paper: tolerance 0.001 & .8641 & .8142 & 337 & 360 & 13.45\\
Tolerance 0 & .8641 & .8142 & 421 & 276 & 14.19\\
Tolerance 0.01 & .8641 & .8142 & 332 & 365 & 13.38\\
Tolerance 0.05 & .8636 & .8115 & 321 & 376 & 13.10\\
Always the section-stack program & .8572 & .8061 & 683 & 0 & 14.55\\
Length by output faces & .8641 & .8142 & 331 & 366 & 13.62\\
\bottomrule
\end{tabular}
\caption{\textbf{Sensitivity of mesh program selection} on the 748 parts
of main-paper Table~1. Fit is the mean of the normalized and
fixed-frame IoU between the replayed solid and the input mesh. Among
candidates whose fit is within the tolerance of the best, the main
paper selects the one with fewer features: search actions, or the
extrusions, revolves, and boxes of a section stack. Ties go to the
search program. The tolerance rows repeat selection over the same
candidates. Always the section-stack program returns the section-stack program
whenever one exists. Length by output faces follows the main rule but
measures length by the face count of each candidate's solid. IND is the
mean with failures scored zero, and $\geq.95$ is the share of the 748
parts with IND at least .95. Stacks and Search count the parts that
return a section-stack or search program, and Features is the mean
feature count of the returned programs.}
\label{tab:supp_selection}
\end{table}

Always returning the section-stack program when one exists gives .857
IND on the 748 parts, against .864 when a search program can win
(Table~\ref{tab:supp_selection}). Its boundary scores are
also close: OSC is .781 against .793 and FeatR95 .719 against .731, and
278 outputs pass the boundary check against 274. Search programs are
shorter: a returned search program has a median of 1 action and 34
output faces, against 10 features and 224 faces for a section stack
(Table~\ref{tab:supp_compact}).
Re-selecting the same candidates with
tolerances from 0 to 0.05 keeps IND between .8636 and .8641. This
tolerance mainly shifts the mix of returned programs: search programs
rise from 276 at 0 to 376 at 0.05, section stacks drop from 421 to 321,
and mean program length falls from 14.19 to 13.10 features.

\paragraph{Surface fitting and cell assembly.}
We measure lengths in units of the longest side of the input bounding
box. We sample points with normals from the mesh and fit analytic
surfaces (planes, cylinders, spheres, cones, and tori) with sequential
RANSAC, using an inlier distance of .003, normal agreement
$|\cos|>.95$, and at least 25 points per connected segment. A curved
surface replaces the best plane only when it has more than 1.02 times as
many inliers, and cones, spheres, and tori require at least 50 inliers.
Each accepted surface is refitted by least squares. Fitting stops after
10 iterations without a new surface, 400 iterations, or 150 surfaces.
Each fitted or predicted surface becomes a carrier: planes form squares
with side length 3.2 times the bounding-box diagonal, cylinders and
cones are cut along their axis to their inliers plus .08 of the
diagonal at each end, and spheres and tori are kept whole. Carriers are
therefore not trimmed to face boundaries. We merge
coincident planes and cylinders and keep the 48 carriers with the most
inliers. We enlarge the bounding box by .03 of the diagonal on each side
and split it with the carriers using Open CASCADE's splitter. When
splitting fails, we add carriers one at a time and skip any carrier that
causes a failure. On the ABC-EVAL meshes, the fitted carriers alone
produce a median of 15 carriers and 138 cells, with 20 inside (90th
percentiles: 29, 715, and 170). A face between two cells is supported
when at least two inlier points from its carrier lie on it. Its vote is
the count of those points whose normals agree with its outward normal
from one cell minus those that disagree. A min-cut then labels the
cells. A supported face with vote $v$ ties the cell on its material side
to the inside terminal and the other cell to the outside terminal, each
with capacity $|v|$. Cutting an unsupported face costs $\lambda$ times
its area, where $\lambda$ is half the point density on supported faces.
Cells touching the enlarged box are tied to the outside. We then fuse
the inside cells and merge faces on the same surface.

\paragraph{Patch candidates.}
We build a patch candidate greedily from a candidate solid. In each round, we take the
connected pieces of the input minus the candidate (to add) and of the
candidate minus the input (to cut), and rank them by volume. For the
largest pieces, we slice each at .5, .2, and .8 of its depth along each
axis and extrude the section as a prism. We keep the prism that most
improves the fit (the mean of the normalized and fixed-frame IoU with
the input mesh) if it gains at least .002. We stop after eight prisms or
once the fit reaches .9985.

\paragraph{Point network.}
The point network has 11,741,911 parameters and takes a cloud of points
with normals as input. Its encoder picks 384 patch centers by farthest
point sampling and groups the 32 nearest points around each center. A
two-stage mini-PointNet embeds each group from point positions relative
to the center and normals, then adds a positional embedding of the
center. A six-layer pre-norm transformer encoder (width 256, eight
heads) mixes the 384 patch tokens. To form per-point features, the
network interpolates the three nearest tokens by inverse distance and
adds a pointwise embedding of position and normal. A six-layer
transformer decoder turns 96 learned slot queries into surface
hypotheses, making predictions after every layer. Each slot predicts an
existence logit, one of six types (plane, cylinder, cone, sphere, torus,
or other), eight parameters and their log-variances, a mask embedding
whose dot product with the per-point features scores point membership,
and an embedding whose pairwise products predict surface adjacency.
During training, the Hungarian algorithm matches slots to true surfaces
using a cost over type probability, mask cross-entropy, mask Dice, and
parameter error (weights 2, 5, 2, 2). Applied at every decoder layer,
the loss sums existence cross-entropy (weight 1, absent slots
down-weighted to 0.2), type cross-entropy (1), mask cross-entropy (5),
mask Dice (2), a smooth-$\ell_1$ parameter loss (5), a Gaussian
likelihood for the log-variances (0.2), and adjacency cross-entropy
(0.5). The training store holds 134,510 Zero-to-CAD parts with 10,000
labeled surface points each, and contains no ABC part. We sample 4,096
points per part at each step; surfaces with at least eight points
become targets, up to the 96 largest. We train for 95,000 steps with
batch 32 and AdamW (learning rate $2\cdot10^{-4}$, weight decay .05,
1,000 warm-up steps, cosine decay to 5\%). Training takes 5.7
GPU-hours, and we keep the checkpoint with the highest surface recall
on 1,286 training-validation parts set aside during training.

\paragraph{Completion network.}
The completion network has 11,829,338 parameters. It extends the point
network with three extra inputs, each scaled by a learned gate
initialized to zero so the model starts as an exact copy of the point
network. Eight per-point field features (whether a fitted surface
explains the point, the distance to it, and that surface's type) enter
both the patch encoder and the pointwise embedding. A small MLP maps
each fitted surface's type and eight parameters to a token appended to
the decoder memory, letting the model predict missing surfaces relative
to known ones. Only surfaces missing from the field serve as targets;
points on known surfaces act as negatives for every slot. Training uses
208,835 Zero-to-CAD and 29,954 ABC training parts, with no ABC-EVAL
part included. We simulate partial fields for each sample: half follow
the measured retention of the analytic fit by surface type and size,
20\% use a uniform keep rate, 25\% keep the largest surfaces first, and
5\% start from an empty field. We also jitter known parameters and
rotate half the samples arbitrarily. At inference, the analytic fit
stays whole. Slots claim only unexplained points, and we refit each
proposal by least squares with support down to eight points for planes
and 16 for curved surfaces. Before building the cells, we drop any
proposal whose residual exceeds twice the fit tolerance.

\paragraph{Candidate-stage face policy.}
In their candidate stage, the mesh and point routes use a second face
policy with 2.75M parameters. It is a 4-layer typed graph transformer
(width 192, 6 heads) operating over the face graphs of the target and
the current solid. Like the B-rep policy, it predicts an action family
with target-entity arguments and never outputs numbers. We train it
only on Zero-to-CAD targets by imitation followed by rejection-sampling
fine-tuning with a KL anchor ($\beta=.02$). It guides a best-first
search (budget 100) on the selected surface reconstruction, and the
resulting program enters candidate selection. The B-rep and drawing
rows do not use this policy.

\begin{table*}[t]
\centering\small
\setlength{\tabcolsep}{3.7pt}
\begin{tabular}{@{}llrrrrrrrr@{}}
\toprule
& & \multicolumn{4}{c}{Face-count tier, mean IND} & & & &\\
\cmidrule(lr){3-6}
Method & Input & 1--15 & 16--40 & 41--100 & $>100$ & Mean IND & Scored & $\#\geq.95$ & $\#\geq.999$ \\
\midrule
\emph{Input mesh returned (reference)} & mesh & -- & -- & -- & -- & .856 & -- & 456 & -- \\
CAD-Recode v1.5~\cite{rukhovich2025cadrecode} & points & .626 & .500 & .423 & .356 & .478 & 375 & 67 & 3 \\
cadrille-RL~\cite{kolodiazhnyi2026cadrille} & points & .706 & .628 & .572 & .468 & .595 & 453 & 76 & 1 \\
CADENA~\cite{kabisov2026cadena} & mesh & .768 & .695 & .629 & .502 & .649 & 430 & 186 & 4 \\
\midrule
Carrier route, no completion, no merging & mesh & .741 & .627 & .520 & .404 & .575 & 378 & 226 & -- \\
Carrier route, no merging & mesh & .755 & .649 & .512 & .402 & .582 & 374 & 236 & 72 \\
\ours Carrier route & mesh & .765 & .689 & .600 & .482 & .635 & 411 & 249 & -- \\
Priority selector & mesh & .816 & .747 & .709 & .640 & .729 & 456 & 294 & -- \\
Restricted pool & mesh & -- & -- & -- & -- & .693 & -- & 209 & -- \\
Section stacks only & mesh & -- & -- & -- & -- & .781 & -- & 373 & -- \\
\ours Full pool & mesh & .872 & .859 & .839 & .764 & .834 & 456 & 404 & 188 \\
\bottomrule
\end{tabular}
\caption{\textbf{Mesh variants by whole-file complexity on ABC-EVAL}
(533 files, first STEP root). The four face-count tiers contain 137,
134, 129, and 133 files. Missing or unscorable outputs are scored as
zero. CAD-Recode and cadrille use 256 samples per mesh with greedy
decoding. Our rows return solids from earlier checkpoints of the mesh
route. ``No merging'' omits guarded face merging; ``no completion''
omits the completion network and uses a retrained point network instead.
The priority selector takes the first usable candidate, starting with
the Carrier route. The Full pool selects by fit to the input mesh, which
provides the reference geometry. Dashes indicate values not reported.}
\label{tab:supp_mesh_detail}
\end{table*}

\paragraph{ABC-EVAL mesh variants.}
Table~\ref{tab:supp_mesh_detail} breaks down the ABC-EVAL mesh
evaluation by complexity tier. Guarded face merging fuses the carrier
cells and merges adjacent faces and edges on the same surface or curve.
It keeps a merge only if it opens no shell, creates no edge shared by
more than two faces, and still yields a solid; this raises IND from
.582 to .635. Programs found on the Full pool's solids reproduce those
solids exactly on 211 parts, but are volume-exact against the reference
on only 61, because input-fit selection returns stacks that approximate
curved faces.

\paragraph{Selectors.}
On the 533 ABC-EVAL files, the priority selector scores .7286 IND, a
selector with structural vetoes and input IoU scores .8136, a
vetoes-only selector scores .7974, a selector with a surface F-score
term scores .7984, and the Full pool reaches .8336. On a 400-part
training-validation set, the priority selector, input-IoU selector,
vetoes-only selector, and Full pool score .9248, .9623, .9495, and
.9707.

\paragraph{Damaged meshes.}
On 200 paired training-validation parts, clean-input mean IND is .963
for the Full pool, .912 for the priority selector, and .846 for the
Carrier route. When the mesh has holes, these means drop to .867, .895,
and .838; under 0.5\% vertex jitter, they fall to .711, .640, and .321.
Holes reduce Full-pool IND by .096 ($[-.124,-.069]$) and
priority-selector IND by .017 ($[-.039,+.005]$). We ran no baseline on
these 200 parts; the next paragraph compares baselines under the same
jitter.

\begin{table}[t]
\centering\footnotesize
\setlength{\tabcolsep}{3pt}
\begin{tabular}{@{}lrrrr@{}}
\toprule
& \multicolumn{2}{c}{Mean IND} & \multicolumn{2}{c}{IND $\geq.95$}\\
\cmidrule(lr){2-3}\cmidrule(l){4-5}
Method & Clean & Jittered & Clean & Jittered\\
\midrule
CAD-Recode v1.5 (points) & .749 & .405 & .21 & .02\\
cadrille-RL (points) & .783 & .667 & .25 & .01\\
CADENA (mesh) & .861 & .667 & .48 & .08\\
\ours \sysname (mesh) & .954 & .537 & .90 & 0\\
\ours \sysname, search only & .775 & .138 & .71 & 0\\
\midrule
\multicolumn{5}{@{}l}{\emph{\sysname (mesh), two changes}}\\
\ours Larger search budget & -- & .545 & -- & 0\\
\ours Noise-aware tolerance & -- & .556 & -- & .01\\
\bottomrule
\end{tabular}
\caption{\textbf{Robustness to 0.5\% vertex jitter} on 100 parts of
main-paper Table~1 (50 ABC-EVAL files, 50 CADBench parts) on which
every method returns a scorable output from the clean input. Every
vertex of the input mesh moves by Gaussian noise with $\sigma$ equal to
0.5\% of the bounding-box diagonal; the point methods receive 256
points sampled from the jittered mesh. Outputs are scored against the
clean mesh; failures score zero. Clean values are the stored runs of
main-paper Table~1 on these parts. Search only: the search program
without the section-stack fallback; its clean value comes from the
search-only row of main-paper Table~1. Bottom block: our route under
jitter with a larger search budget ($V=400$, wall limits doubled) or
with a noise-aware section tolerance. --: not measured.}
\label{tab:supp_noise}
\end{table}

\paragraph{Sensitivity to mesh noise.}
Our mesh route is built for clean geometry. In
Table~\ref{tab:supp_noise}, we apply the same 0.5\% vertex jitter to
100 parts from main-paper Table~1 (50 ABC-EVAL files and 50 CADBench
parts) where every method returns a scorable output on the clean input.
Our route suffers the largest drop: mean IND falls from .954 to .537
($-.4162$ $[-.4773,-.3549]$, paired bootstrap 95\% interval), and the
share of parts with IND of at least .95 falls from .90 to 0. By
comparison, cadrille-RL drops from .783 to .667 ($-.1164$
$[-.1415,-.0915]$), CADENA drops from .861 to .667 ($-.1939$
$[-.2438,-.1469]$), and CAD-Recode v1.5 drops from .749 to .405
($-.3448$ $[-.4073,-.2850]$). Under jitter, CAD-Recode fails on 25
parts and CADENA fails on 6, while cadrille-RL and our route fail on
none. Our route drops from .914 to .540 on the ABC-EVAL files and from
.993 to .535 on the CADBench parts. Search alone falls from .775 to
.138 and returns no program on 60 of the 100 parts, leaving the
section-stack fallback to hold the route at .537. On training-side
parts, jitter breaks the reconstructed solid into many faces (a median
of 38 faces on clean meshes versus 936 on jittered ones), and search
rarely succeeds on solids with more than 500 faces. Two changes help
slightly. A larger search budget ($V=400$, wall limits doubled) raises
mean IND to .545 ($+.008$ $[+.000,+.019]$), and a noise-aware section
tolerance raises it to .556 ($+.018$ $[+.003,+.036]$). This tolerance
simplifies each section profile in proportion to a noise level
estimated from the input mesh; it turns off on meshes judged
noise-free, so clean results stay unchanged. With each change,
search alone reaches .176 and .186, respectively, and returns no program
on 52 and 49 parts. Even then, our route stays below cadrille-RL and CADENA under
jitter. Because the clean scores come from the stored runs of
main-paper Table~1, each drop also includes run-to-run variation.
Because these 100 parts exclude clean failures, every clean mean
exceeds its main-paper Table~1 value.
CADBench's own noisy-mesh track is milder than this jitter. Following
its published protocol with its released generator (isotropic
remeshing, Gaussian vertex noise with $\sigma=0.005$ in the $[-1,1]^3$
frame, then one Laplacian smoothing step), we perturb 96 CADBench
parts, 16 per split and tier. Our route reaches a mean CADBench IoU of
.806 and a median of .923, against a mean of .944 on the clean meshes
of the same parts, and its mean IND is .764. Search alone returns a
program on only 19 of the 96 parts. On its full noisy track, CADBench
reports IoU drops of .051 for CADFit, .108 for cadrille, .162 for
CAD-Recode and .577 for CADEvolve; our drop in mean IoU, .138, is measured on a
different and smaller sample, so this comparison is only indicative.
At the generator's default setting ($\sigma=0.001$, no smoothing), the
route reaches a mean of .819 and a median of .937.

\paragraph{Oriented points.}
On 33 training-validation parts, the Full pool scores .974 on clean
meshes and .832 on Poisson meshes built from 8,192 oriented samples;
the Carrier route scores .834 on meshes and .835 directly on the same
points, and the priority selector scores .951. The number of parts
where the Full pool reaches IND $\geq.99$ drops from 28 to 1 of 33.

\paragraph{Stage timing.}
On 477 ABC-EVAL meshes, the Full pool's candidate stage takes a median
of 153.9 worker-seconds per part (mean 277.6, 90th percentile 718.4).
The Carrier route takes 58.2\,s without completion and 102.5\,s with
it. These times exclude upstream preprocessing and export; policy
search and pool selection have wall limits of 600 and 1,500\,s.

\section{Drawings}
\label{sec:supp_drawings}

\begin{table}[t]
\centering\footnotesize
\setlength{\tabcolsep}{3pt}
\begin{tabular}{@{}lrrr@{}}
\toprule
& DeepCAD & Fusion~360 & Manual\\
\midrule
Ortho2CAD SFT (8B)~\cite{joglekar2026ortho2cad} & .792 & .370 & .193\\
Ortho2CAD RL (8B)~\cite{joglekar2026ortho2cad} & .776 & .560 & .300\\
GPT~5.5 ITS~\cite{joglekar2026ortho2cad} & .846 & .817 & \textbf{.750}\\
\midrule
\multicolumn{4}{@{}l}{\emph{Ours: program}}\\
\ours Carve program (not learned) & .885 & .823 & .506\\
\ours + learned correction & \textbf{.893} & \textbf{.831} & .506\\
\bottomrule
\end{tabular}
\caption{\textbf{Reconstruction from orthographic drawings on the
Ortho2CAD test sets} (100 drawings each). Entries report mean IoU after
CAD-Coder normalization and alignment~\cite{doris2025cadcoder}, with
failures scored as zero, as published by Ortho2CAD. Manual refers to
manually dimensioned sheets. GPT~5.5 ITS is GPT~5.5 with Ortho2CAD's
inference-time scaling~\cite{joglekar2026ortho2cad}. Because the
published values~\cite{joglekar2026ortho2cad} do not include per-part
scores, differences are not tested. Every test program replays to its
scored solid. The correction is not applied to manual sheets.}
\label{tab:supp_drawings}
\end{table}

\paragraph{Pipeline.}
Each input is a raster sheet containing front, top, and right views. A
classical vectorizer reads printed dimensions, scale, and extents; when
it fails (all 100 manually dimensioned sheets), we build a frame from
the views. We trace outlines from the raster using flood fill, contour
extraction, and polygon simplification, with optional symmetry and axis
snapping. The carve intersects a box from the extents with each view's
outline prism (or a revolved base). Each cut candidate is a face of one
view's line arrangement over a depth interval whose ends come from
axis-aligned lines in the other two views or from the material extreme.
Greedy best-first search keeps a cut when the re-rendered visible and
hidden lines agree better with the sheet. Next, the 1.67M-parameter
correction network predicts occupancy on a $96^3$ grid. Each region where
it disagrees with the carve (regions thicker than one voxel with at
least 27 voxels, at most 24 per part) turns into axis-aligned boxes written as one
fuse or cut. If the total change is below 5\% of the volume, we keep the
carve unchanged.
The correction network has 1,669,002 parameters and learns a residual on
top of the carve. A 2D U-Net shared across the three views (four levels,
widths 16 to 128) takes 15 channels per view at $96\times96$: visible
lines, hidden lines, the silhouette, a three-channel view code, two log aspect ratios,
two coordinate channels, and five channels comparing the sheet with the
re-rendered lines of the carve. Its features are averaged over each
closed region in the view and back-projected into the $96^3$ grid along
with the visual hull and the carve. A 3D U-Net (four levels, widths 16
to 96) averages features over the cells formed by those back-projected
regions and adds a zero-initialized residual to the carve's occupancy
logit, so training starts from the carve itself. The loss is voxel
cross-entropy, given three times the weight near the true surface, plus
a soft IoU term. We train for 209,500 steps with batch size 3 and AdamW
(learning rate $5\cdot10^{-4}$, weight decay $10^{-4}$, one-cycle
schedule), taking about 8.8 GPU-hours. We keep the checkpoint with the
best validation IoU on training-side parts. The 139,618 training sheets
come from 124,792 Zero-to-CAD parts and the training splits of DeepCAD
(8,727) and Fusion~360 (6,099), drawn in Ortho2CAD's sheet convention.
Dropped line pixels, missing chunks, and speckle imitate imperfect line
images, and in 10\% of samples the visual hull replaces the carve as
input. Two of the 348 CADBench parts share an ABC identifier with these
DeepCAD training parts.

\paragraph{Results.}
On the Ortho2CAD test sets (Table~\ref{tab:supp_drawings}), the carve
program without learned parameters outperforms both 8B Ortho2CAD models.
With the learned correction, it scores .893 on DeepCAD and .831 on
Fusion~360, ahead of GPT~5.5 ITS, Ortho2CAD's inference-time
scaling~\cite{joglekar2026ortho2cad} (.846 and .817). On manually
dimensioned sheets, however, it trails at .506 versus .750. Every
program is a stand-alone CadQuery script that replays to its scored
solid in a separate process, with a symmetric-difference volume of at
most $10^{-5}$ of the volume (100 of 100 in every set and
configuration). Of the 300 programs, 48 reach IoU .999.
No part in the three Ortho2CAD test sets shares a file identifier with
our training data, and no Fusion~360 test part shares a design with it.
We do not use Ortho2CAD's Zero-to-CAD test set because it overlaps our
Zero-to-CAD test split.
In the main paper's 748-part comparison, the two routes take different
sheets. Ortho2CAD receives sheets from its own released generator,
checked for parity against its published test images; these sheets print
the bounding-box dimensions. Our route receives our own three-view
sheets in the same layout, drawn from the reference STEP with hidden
lines and no dimension text, so it builds its frame from the views. On
ABC-EVAL, our route produces a program for 487 of 490 parts with a
drawing record.

\begin{table*}[t]
\centering\small
\setlength{\tabcolsep}{4pt}
\begin{tabular}{@{}llrrrrrrrr@{}}
\toprule
& & & & & & & & \multicolumn{2}{c}{348 CADBench parts}\\
\cmidrule(l){9-10}
Method & Sheets & IND & $\geq.95$ & OSC & FeatR95 & Verified & Exact & IoU & SIoU\\
\midrule
Ortho2CAD RL~\cite{joglekar2026ortho2cad} & Ortho2CAD's generator & .334 & .008 & .069 & .005 & 1 & 0 & .443 & .323\\
\ours \sysname & ours (main-paper Table~1) & .744 & .349 & .449 & .132 & 0 & 0 & .799 & .662\\
\ours \sysname & Ortho2CAD's generator & .725 & .341 & .443 & .173 & 12 & 2 & .768 & .633\\
\bottomrule
\end{tabular}
\caption{\textbf{Drawing route on the sheets of Ortho2CAD's own
generator} for the 748 parts of main-paper Table~1. Ortho2CAD's
released generator prints the bounding-box dimensions on its sheets;
our own sheets have the same layout, with hidden lines and without
dimension text. Our route is the carve program with the learned
correction, as in main-paper Table~1. Columns, eligibility and scoring
as in main-paper Table~1; failures score zero. The generator gives no
sheet for 30 parts, which score zero for both methods. The first two
rows are those of main-paper Table~1.}
\label{tab:supp_o2csheets}
\end{table*}

\paragraph{Ortho2CAD's sheets on the 748 parts.}
In Table~\ref{tab:supp_o2csheets}, we evaluate our route on sheets from
Ortho2CAD's own generator so that both methods see the same input. Our
route scores a mean IND of .725 versus .334 for Ortho2CAD RL. The paired
difference is $+.3905$ $[+.3710,+.4102]$ (95\% paired bootstrap
interval, 10,000 resamples), with our route scoring higher on 670 parts,
lower on 33, and tying on 45. By split, the difference is $+.3842$ on
the 400 ABC-EVAL files and $+.3978$ on the 348 CADBench parts. Compared
with our own sheets, mean IND drops by .019 ($-.0188$
$[-.0300,-.0078]$): by .008 on the ABC-EVAL files ($[-.022,+.006]$) and
by .031 on the CADBench parts ($[-.049,-.014]$). Ortho2CAD's generator
produces no sheet for 30 parts, 27 of which have no face in the first
STEP root. Without the learned correction, the carve program scores .662
IND on Ortho2CAD's sheets.

\paragraph{Learned correction on the 748 parts.}
On the 748 parts of main-paper Table~1, adding the 1.67M-parameter
correction network to the carve program raises mean IND from .615 to
.689 on the 400 ABC-EVAL files and from .748 to .807 on the 348 CADBench
parts. The share of parts reaching an IND of at least .95 increases from
.228 to .292 on ABC-EVAL and from .385 to .414 on CADBench. By contrast,
OSC and FeatR95 change very little: OSC moves from .441 to .448 and from
.446 to .450, while FeatR95 moves from .126 to .125 and from .142 to
.139.

\section{Boundary metrics}
\label{sec:supp_metrics}

\paragraph{Oriented surface coverage.}
We define a source \emph{patch} as a maximal set of adjacent faces on
the same underlying surface, merged by Open CASCADE same-domain
unification; disconnected faces on one plane remain separate patches. We
draw about 64 area-uniform samples on each patch's exact surface and
test them against the trim. A sample counts as covered if the nearest
point on an output face lies within $\tau D_S$ and the outward normals
there differ by at most $30^\circ$. Here $D_S$ is the source diameter,
the largest distance between convex-hull vertices of the source
tessellation (median .94 of the bounding-box diagonal). For mesh
outputs, we filter triangles by the normal cone first. Patch coverage is
the area-weighted share of covered samples. OSC takes the unweighted
mean across patches, so a small chamfer carries the same weight as a
large face, and then averages across parts. Area recall and precision
use the same matching. We report $\tau=10^{-3}$. As sensitivity checks,
we test $20^\circ$ and $45^\circ$ cones as well as a thickness-capped
tolerance (distance limited to 10\% of the local wall thickness or gap,
with a floor of $10^{-4}D_S$); the $45^\circ$ cone detects missing
$45^\circ$ chamfers less readily. Raising sample counts from about 64
per patch and 4,096 per output to 256 and 16,384 changes every mean by
at most .002 across 61 pairs.

\begin{table}[t]
\centering\footnotesize
\setlength{\tabcolsep}{3pt}
\begin{tabular}{@{}llrrr@{}}
\toprule
& & \multicolumn{2}{c}{OSC at $\tau$} & FeatR95\\
\cmidrule(lr){3-4}\cmidrule(lr){5-5}
Input & Method & $10^{-3}$ & $10^{-2}$ & $10^{-2}$\\
\midrule
Points & CAD-Recode v1.5 & .189 & .401 & .202\\
& cadrille-RL & .208 & .441 & .224\\
\ours & Ours, 8,192 oriented pts & \textbf{.322} & \textbf{.582} & \textbf{.517}\\
\midrule
Mesh & CADENA & .392 & .612 & .512\\
& StepCAD & .213 & .385 & .202\\
& CADFit & .445 & .504 & .477\\
\ours & Ours & \textbf{.793} & \textbf{.832} & \textbf{.761}\\
\midrule
\ours B-rep & Ours & .796 & .819 & .829\\
\midrule
Drawings & Ortho2CAD RL & .069 & .169 & .034\\
\ours & Ours & \textbf{.448} & \textbf{.729} & \textbf{.515}\\
\bottomrule
\end{tabular}
\caption{\textbf{Surface scores at a looser tolerance} on the 748 parts
of main-paper Table~1, computed from the same stored samples. The
$10^{-3}$ column matches main-paper Table~1 to within .001, except for
the B-rep and point rows, which are scored here on 705 and 651 parts
rather than 712 and 656 (.796 and .322 against .789 and .320). FeatR95
here averages over parts with a scored output.}
\label{tab:supp_osc_tau}
\end{table}

\paragraph{A looser tolerance.}
A tolerance of $10^{-3}D_S$ is strict for methods that approximate a
surface instead of reproducing it. In Table~\ref{tab:supp_osc_tau}, we
therefore recompute OSC and FeatR95 for the rows of Table~1 in the main
paper at $\tau=10^{-2}$. Every method improves, most of all those
whose outputs approximate the surface: our drawing and point routes
gain .28 and .26, and CADENA, cadrille-RL, CAD-Recode and StepCAD gain
.17 to .23 (CADENA's OSC rises from .392 to .612). For each input, our
route has the highest OSC at $10^{-2}$ and the highest FeatR95 over
parts with a scored output; at $10^{-3}$, cadrille-RL leads our point
route on FeatR95 in main-paper Table~1 (.064 against .058). On meshes, it scores .832 and .761
versus .612 and .512 for CADENA; on points, .582 and .517 versus .441
and .224 for cadrille-RL; and on drawings, .729 and .515 versus .169 and
.034 for Ortho2CAD RL. Gains are smallest where outputs mostly reproduce
the source surfaces: .04 for our mesh route, .02 for our B-rep route and
.06 for CADFit.
Aligning the outputs does not close the gap at $10^{-3}$. When we align
each output to its reference with a similarity ICP (scale, rotation and
translation) before scoring, OSC at $10^{-3}$ rises for no method and
falls by up to .04; keeping the better of the aligned and unaligned
score for each part raises it by at most .025. The estimated
misplacement is small: a median scale within .002 of 1, a rotation below
$0.11^\circ$ and a shift below .002 of the source diameter, except for
Ortho2CAD RL (.990, $0.39^\circ$ and .013). These figures cover 672 of the
705 parts of Table~\ref{tab:supp_osc_tau}, those on which every row's
alignment finished.

\paragraph{Boundary check.}
An output passes the boundary check if every source sample has an output
sample within tolerance, every output sample has a source sample within
tolerance in the reverse direction, and, for a B-rep output, it contains
at least one solid. The check does not require volume exactness. We
exclude sources without a solid for every method (48 of the 533 ABC-EVAL
files). We take the first STEP root as the reference and score
multi-solid shapes on the union of their boundaries.
Table~\ref{tab:supp_verified} lists how each count uses the check.

\paragraph{Frame and eligible parts.}
OSC, FeatR95, and the boundary check evaluate each output in the
reference's own STEP frame, that of its first root. The scorer does not
align or rescale outputs. Outputs from CAD-Recode v1.5, cadrille-RL,
CADENA, StepCAD, CADFit, and our B-rep route already lie in that frame;
CADFit and StepCAD export their solids in it. When a method writes in a
different frame, as our mesh route and Ortho2CAD do, we map its output
into the reference frame with a scale and translation before scoring.
Main-paper Table~1 averages OSC and FeatR95 over each row's eligible
parts: those where the reference's first root has a solid and scoring
for that row did not fail or time out. Missing outputs score zero.
FeatR95 also requires at least one recognized feature on the reference.
The caption of main-paper Table~1 calls these the parts with a valid
reference; they also exclude, for each row, parts whose scoring failed
or timed out.

\begin{table}[t]
\centering\small
\setlength{\tabcolsep}{3pt}
\begin{tabular}{@{}p{.42\columnwidth}p{.52\columnwidth}@{}}
\toprule
Count & Definition\\
\midrule
Verified, main-paper Table~1 & returned output passes the check, exact or not\\
Boundary-verified (Gallery) & volume-exact on fresh replay and passes the check\\
Verified$_{100}$, Table~\ref{tab:supp_brep_full} & volume-exact and passes the check against the whole file\\
\bottomrule
\end{tabular}
\caption{\textbf{How each table counts boundary verification}
($\tau=10^{-3}$, $30^\circ$).}
\label{tab:supp_verified}
\end{table}

\paragraph{FeatR95.}
Quiddity~\cite{quiddity} is an open-source feature recognizer for STEP
and B-rep solids distributed as a Python package; we run version 0.5.0
unmodified with default options. Using geometry alone, it labels
features such as holes and hole patterns, fillets, blends, chamfers,
slots, pockets, bosses, steps, and freeform surfaces, and lists the
faces that support each one. Its recall is incomplete: the project's
issue tracker reports that 16.8\% of labeled feature instances on its
benchmark receive no face, and on 500 of our sources, 35.5\% of faces
belong to no recognized feature. Because its tolerances are absolute, we
run it on a copy of the source scaled to diameter 100 and keep the
result only if the face counts match. Quiddity labels only source faces;
it never reads the output, and neither OSC nor the boundary check uses
it. We count identical support face sets once and drop supports with a
face-index mismatch. A feature is preserved when the minimum coverage
across its supporting faces is at least .95, an operational threshold.
FeatR95 is the share of preserved features per part, averaged over parts
with at least one recognized feature; missing outputs score zero. Faces
with no recognized feature do not enter FeatR95, but OSC still covers
them.

\paragraph{Sensitivity to the source representation.}
Re-expressing 29 sources as NURBS leaves their patch counts unchanged
and shifts OSC by .0008 on average (at most .008), whereas it alters the
recognized supports on all 29 sources and shifts FeatR95 by .12 on
average. OSC is therefore the stable boundary metric, while FeatR95 is a
descriptive feature-level indicator.

\paragraph{Small features.}
Volume overlap barely reflects small boundary losses. Removing a 0.2\,mm
chamfer keeps volume IoU at .99994 while dropping OSC to .704. In a
training-side study, 12 of 254 volume-exact outputs fail the boundary
check at $\tau=10^{-3}$, and 16 fail at $\tau=10^{-4}$.

\begin{table}[t]
\centering\footnotesize
\setlength{\tabcolsep}{3pt}
\begin{tabular}{@{}lrrrr@{}}
\toprule
& Volume & \multicolumn{3}{c}{OSC at $\tau$}\\
\cmidrule(lr){3-5}
Defect of the output & IoU & $10^{-4}$ & $10^{-3}$ & $10^{-2}$\\
\midrule
Pocket missing & .9583 & .815 & .815 & .815\\
One of four holes missing & .9910 & .963 & .963 & .963\\
One of 12 notches missing & .9933 & .940 & .940 & .940\\
Shaft groove missing & .9971 & .667 & .667 & .778\\
Four 3\,mm fillets missing & .9987 & .741 & .832 & .943\\
$5^\circ$ wall draft missing & .8945 & .336 & .359 & .523\\
Scaled by 1.01 & .9649 & .003 & .263 & 1.000\\
Solid block for a thin open box & .0764 & .545 & .545 & .545\\
\bottomrule
\end{tabular}
\caption{\textbf{Constructed defects.} Volume IoU and OSC of an output
with one known defect against its source, at three tolerances $\tau$
(fractions of the source diameter). The main paper reports OSC at
$\tau=10^{-3}$.}
\label{tab:supp_osc_panel}
\end{table}

\paragraph{Constructed defects.}
We also score 28 constructed source-output pairs with known differences.
A source scores 1 against itself, a fine triangle mesh of the same solid
scores 1, two touching solids score 1 against their fused block, and a
rigid motion applied to both shapes leaves all scores unchanged.
Splitting source faces into pieces changes neither the patch count nor
any score. Table~\ref{tab:supp_osc_panel} shows outputs with one defect.
Because OSC averages over patches, a missing feature costs its share of
source patches regardless of volume: a missing groove keeps volume IoU
at .9971 but reduces OSC to .667. Missing fillets retain partial credit
that increases with $\tau$. Meshes with noisy or staircase facets fail
the normal cone at every $\tau$.

\begin{table}[t]
\centering\footnotesize
\setlength{\tabcolsep}{3pt}
\begin{tabular}{@{}lrrrrr@{}}
\toprule
& Mean & \multicolumn{3}{c}{OSC at $\tau$} & Passing\\
\cmidrule(lr){3-5}
Route & IoU & $10^{-4}$ & $10^{-3}$ & $10^{-2}$ & check\\
\midrule
Full pool, mesh & .970 & .777 & .831 & .859 & 181\\
CADENA~\cite{kabisov2026cadena}, mesh & .854 & .295 & .510 & .698 & 33\\
Policy search, B-rep & .890 & .789 & .794 & .809 & 268\\
B-rep + mesh, rule selector & .970 & .833 & .871 & .894 & 284\\
B-rep + mesh, OSC selector & .974 & .872 & .902 & .919 & 285\\
\bottomrule
\end{tabular}
\caption{\textbf{Boundary fidelity on 464 training-validation parts.}
OSC at three tolerances $\tau$; the boundary check uses
$\tau=10^{-3}$. The rule selector takes the B-rep
policy's solid if it is volume-exact and the Full pool's otherwise; the
OSC selector scores both against the input B-rep. The policy rows
include 26 evaluation errors or crashes and 30 time caps.}
\label{tab:supp_boundary_tvl}
\end{table}

\paragraph{Mesh and B-rep boundaries.}
Across the 322 parts in Table~\ref{tab:supp_boundary_tvl} where both
routes reach an IoU of at least .99, the B-rep route improves OSC by
.0638 $[.0473,.0809]$ over the Full pool; area recall increases from
.9914 to .9979, and precision increases from .9902 to .9979. Across all
464 parts, the share of preserved features for the Full pool, B-rep
policy, and CADENA is .58, .56, and .09 for chamfers; .71, .74, and .21
for holes; and .68, .74, and .14 for fillets. Moving from
$\tau=10^{-4}$ to $10^{-2}$, OSC rises from .777 to .859 for the Full
pool, from .789 to .809 for the B-rep policy, and from .295 to .698 for
CADENA, which stays lowest at every tolerance.

\begin{table}[t]
\centering\small
\setlength{\tabcolsep}{3pt}
\begin{tabular}{@{}lrrr@{}}
\toprule
Output & Eligible & OSC & Passing check\\
\midrule
CAD-Recode v1.5 & $420 \to 395$ & $.188 \to .243$ & $4 \to 9$\\
cadrille-RL & $399 \to 372$ & $.207 \to .260$ & $3 \to 6$\\
\bottomrule
\end{tabular}
\caption{\textbf{First-root boundary reference on ABC-EVAL.}
Comparison of whole-file and first-root scores at $\tau=10^{-3}$ on the
same available rows. Eligible counts exclude sources with no usable solids.}
\label{tab:supp_abc_boundary}
\end{table}

\paragraph{Boundary reference on ABC-EVAL.}
Scoring against the first STEP root rather than the whole file raises
OSC to .243 for CAD-Recode and .260 for cadrille on the available rows
(Table~\ref{tab:supp_abc_boundary}). Across 372 paired first-root rows,
mean cadrille minus CAD-Recode is .0091 (95\% CI $[-.0055,.0236]$).

\section{ABC-EVAL protocol and external baselines}
\label{sec:supp_abc}

\begin{table}[t]
\centering\small
\begin{tabular}{@{}lrrr@{}}
\toprule
Method & $n$ & Mean IND & Scored\\
\midrule
CADFit~\cite{nehme2026cadfit} & 148 & .454 & 82\\
Carrier route & 148 & .554 & 103\\
Full pool & 148 & .842 & 127\\
\bottomrule
\end{tabular}
\caption{\textbf{CADFit on a 148-file ABC-EVAL subset.}
The subset contains 37 files per face-count tier. All methods are evaluated on
the same files. IND counts missing and unscorable outputs as zero.}
\label{tab:supp_abc_subsets}
\end{table}

\paragraph{Test sets, tiers and selection.}
ABC-EVAL contains 533 ABC files~\cite{koch2019abc} drawn from the ABC
parts of MM-CAD~\cite{bharathi2026mmcad}. We ordered candidates by a
fixed hash within four tiers of whole-file B-rep face count (1--15,
16--40, 41--100 and more than 100 faces) and took the same number from
each tier. Removing files that lack any of MM-CAD's modalities without
replacement leaves 137, 134, 129 and 133 files. None of its files appears in our
training data (checked by identifier). The 748 parts in main-paper
Table~1 combine the 400 single-solid files from ABC-EVAL with 348
single-solid CADBench ABC parts, 58 per split and tier. This size lets
every method run on every part, including per-part search baselines
such as CADFit (mean 1,804\,s per part). CADBench divides each split's
3,000 designs into its own low, medium and high face-count tiers of
1,000 designs, with boundaries set per split~\cite{doris2026cadbench}.
Main-paper Fig.~3 uses the ABC-EVAL tiers, while main-paper Table~3
and Table~\ref{tab:supp_cadbench} use the CADBench tiers. On meshes,
we return the shortest candidate whose fit to the input mesh lies
within 0.001 of the best (Sec.~\ref{sec:supp_mesh}, Mesh programs).

\paragraph{First STEP root.}
Every method and check loads each file with the CadQuery STEP importer
and takes the first transferred root, which is the first shape returned
by Open CASCADE's STEP reader. That root is the whole file in 367 of
the 533 files. Among the 166 files with several roots, the first root
is a single solid in 123, a compound in 36 and a shell in 7; 29 first
roots have no face and score zero for every method. Out of the 533
files, 400 contain one solid, 121 contain several and 12 contain none.
Mesh inputs exist for 477 files, and the IND scorer rejects another 21
non-manifold reference meshes.

\paragraph{Failures, reference ceiling and paired intervals.}
For each row of main-paper Table~1, Table~\ref{tab:supp_validonly}
lists how many parts lack a valid output along with the mean IND over
valid outputs alone. Valid-only means favor rows that fail often.
From the B-rep input, 403 of the 748 parts receive a volume-exact
program. Because the B-rep route has no section-stack fallback and its
143 failures score zero, its overall mean IND (.736) falls below the
mesh route's (.864), even though it produces about four times as many
volume-exact programs (403 against 102). Over valid outputs, the B-rep
route scores .910 (Table~\ref{tab:supp_validonly}). Since the mesh
route's input is a tessellation of the same B-rep, the stored outputs of
the two routes can also be combined: returning the B-rep program where it is volume-exact within $V=100$
(403 parts) and the mesh route's output elsewhere gives mean IND .870,
a share of .825 with IND of at least .95, OSC .826, FeatR95 .792 and
407 verified outputs. The mesh route alone gives .864, .814, .793, .731
and 274, and the B-rep row alone has 417 verified outputs.
CADFit fails on 347 parts: 185 from ABC-EVAL and 162 from CADBench.
These failures include 47 ABC-EVAL files with no usable input mesh
(StepCAD has the same 47), 41 ABC-EVAL time-outs at the 7,200\,s cap
and 25 invalid B-reps. On CADBench, 76 of 346 runs hit the 7,200\,s
cap.
CADBench reports a valid-shape rate of .981 for CADFit across its
splits~\cite{doris2026cadbench}. The two evaluations differ as follows.
We ran CADFit's public release on the input meshes of main-paper
Table~1 with a 7,200\,s per-part cap, and main-paper Table~1 reports
mean IND with every failure scored zero. CADBench instead reports tier
medians; it also scores failures zero, but a median changes little
while failures are few.
Returning the reference B-rep itself scores .932 IND across the 748
parts (.890 on ABC-EVAL and .980 on CADBench): every scorable part
reaches at least .999, while 51 references cannot be scored.
Paired bootstrap 95\% intervals (10,000 resamples) give $+.193$
$[+.168,+.218]$ IND for the mesh route over CADENA (better on 640
parts, worse on 48) and $+.409$ $[+.390,+.429]$ for the drawing route
over Ortho2CAD RL (better on 677, worse on 28).

\begin{table}[t]
\centering\footnotesize
\setlength{\tabcolsep}{3pt}
\begin{tabular}{@{}lrrr@{}}
\toprule
Main-paper Table~1 row & IND & Failures & Valid-only IND\\
\midrule
CAD-Recode v1.5 (points) & .482 & 212 & .673\\
cadrille-RL (points) & .641 & 59 & .696\\
\ours \sysname (points) & .609 & 155 & .768\\
CADENA (mesh) & .672 & 131 & .814\\
StepCAD (mesh) & .546 & 101 & .631\\
CADFit (mesh) & .444 & 347 & .827\\
\ours \sysname, search only (mesh) & .648 & 182 & .857\\
\ours \sysname (mesh) & .864 & 64 & .945\\
Ortho2CAD RL (drawing) & .334 & 48 & .357\\
\ours \sysname (drawing) & .744 & 54 & .801\\
\ours \sysname (B-rep) & .736 & 143 & .910\\
\bottomrule
\end{tabular}
\caption{\textbf{Failures and valid-only means} on the 748 parts of
main-paper Table~1. A failure has no output, an invalid output or an
unresolvable IoU; it scores zero in the main table. For Ortho2CAD RL,
26 further valid outputs score IoU 0; counting them as failures gives
74 failures and .371.}
\label{tab:supp_validonly}
\end{table}

\begin{table}[t]
\centering\footnotesize
\setlength{\tabcolsep}{3pt}
\begin{tabular}{@{}lrrr@{}}
\toprule
& IND & IND & $\geq.95$\\
Main-paper Table~1 row & 748 & 697 & 697\\
\midrule
CAD-Recode v1.5 (points) & .482 & .511 & .118\\
cadrille-RL (points) & .641 & .681 & .144\\
\ours \sysname (points) & .609 & .643 & .264\\
CADENA (mesh) & .672 & .713 & .354\\
StepCAD (mesh) & .546 & .581 & .146\\
CADFit (mesh) & .444 & .474 & .323\\
\ours \sysname, search only (mesh) & .648 & .692 & .634\\
\ours \sysname (mesh) & .864 & .916 & .864\\
Ortho2CAD RL (drawing) & .334 & .355 & .009\\
\ours \sysname (drawing) & .744 & .789 & .372\\
\ours \sysname (B-rep) & .736 & .788 & .722\\
\midrule
\emph{Reference B-rep returned} & \emph{.932} & \emph{1.000} & \emph{1.000}\\
\bottomrule
\end{tabular}
\caption{\textbf{Main-paper Table~1 on the parts with a scorable
reference.} Mean IND over all 748 parts, mean IND over the 697 parts
whose reference can be scored, and the share of those 697 parts with IND
at least .95. Failures score zero. Outputs are scored against the
reference mesh. The last row returns the reference B-rep itself,
tessellated and scored in the same way. On 51 parts it produces no valid
IoU, so their references cannot be scored. The 697 column drops these 51
parts from every row. The 748 column keeps them in every row: a row
scores zero on those parts unless its output has a valid IoU against the
reference mesh. Six rows have a valid IoU on eight or nine of the 51
parts (\sysname from meshes, drawings, and points; CADENA; cadrille-RL;
and Ortho2CAD RL). In Table~\ref{tab:supp_validonly}, a row's failure
count includes only its own missing or invalid outputs and unresolvable
IoUs, so it can be below 51; Ortho2CAD RL has 48.}
\label{tab:supp_scorable}
\end{table}

Restricting evaluation to the 697 parts with a scorable reference
raises every mean without changing the row ranking
(Table~\ref{tab:supp_scorable}). On this subset, the mesh route
reaches .916 IND compared with .713 for CADENA, and the drawing route
reaches .789 compared with .355 for Ortho2CAD RL.

\begin{table*}[t]
\centering\small
\setlength{\tabcolsep}{5pt}
\begin{tabular}{@{}lrrrrr@{}}
\toprule
Returned programs & $n$ & Tokens & Operations & Lines & Faces\\
\midrule
\multicolumn{6}{@{}l}{\emph{748 parts of main-paper Table~1}}\\
CAD-Recode v1.5 (points) & 701 & 285 & 30 & 3 & 21\\
cadrille-RL (points) & 701 & 163 & 19 & 3 & 16\\
\ours \sysname (points) & 604 & 19,426 & 1,405 & 7 & 2,388\\
CADENA (mesh)$^b$ & 337 & 140 & 3 & 16 & 23.5\\
StepCAD (mesh) & 682 & 493 & 43 & 7 & 29\\
CADFit (mesh) & 531 & 1,076 & 87 & 87 & 46\\
\ours \sysname, search only (mesh) & -- & -- & -- & -- & 53\\
\ours \sysname (mesh): search programs & 360 & -- & -- & -- & 34\\
\ours \sysname (mesh): section stacks$^a$ & 701 & 1,114 & 101 & 7 & 224\\
Ortho2CAD RL (drawing) & 718 & 370 & 44 & 18 & 14\\
\ours \sysname (drawing) & 710 & 6,449 & 22 & 30 & 578\\
\ours \sysname (B-rep) & 529 & 656 & 51 & 69 & 37\\
\midrule
\multicolumn{6}{@{}l}{\emph{CADBench, ours}}\\
\ours STEP input, DeepCAD & 2,969 & 220 & 18 & 29 & 10\\
\ours STEP input, Fusion~360 & 2,697 & 374 & 23 & 41 & 12\\
\ours STEP input, ABC Extrude & 2,757 & 995 & 104 & 115 & 71\\
\ours STEP input, ABC All-Ops & 2,355 & 870 & 68 & 86 & 40\\
\ours Mesh input, search only (348) & -- & -- & -- & -- & 64.5\\
\ours Mesh input, with fallback (348) & -- & -- & -- & -- & 74\\
\bottomrule
\end{tabular}
\caption{\textbf{Program size.} Each row reports medians per returned
program: tokens, CadQuery operations, and non-empty script lines
measured with CADBench's code metrics~\cite{doris2026cadbench}, along
with the faces of the output solid. Here $n$ is the number of measured
programs. Faces are counted over the output solids, and this count can
differ from $n$. Operations count every CadQuery call, including sketch
segments and selectors. Line counts depend on formatting: CAD-Recode and
cadrille write one chained expression. For the mesh route, programs are
split into search programs and section stacks. $^a$Script counts are
over the section-stack programs written for 701 parts; face counts are
over the 337 returned. $^b$ABC-EVAL parts only. --: not available; for
mesh-route search programs, the text reports action counts instead.}
\label{tab:supp_compact}
\end{table*}

Our search programs are lists of actions on the target's faces, which
the emitter writes out as scripts (Table~\ref{tab:supp_cqemit}).
Returned search programs from the mesh route have a median of 1 action,
while those from the B-rep route have a median of 3 actions across the
668 programs that build a solid. Section-stack programs are long
because each section is its own profile (Table~\ref{tab:supp_compact}).

\paragraph{CADFit on a 148-file subset.}
The CADFit comparison uses 37 files per face-count tier
(Table~\ref{tab:supp_abc_subsets}). CADFit~\cite{nehme2026cadfit}
optimizes extrude, revolve, fillet and chamfer programs for IoU against
the input mesh using a DINOv2-based sketch prior. On this subset,
CADFit produces 82 scorable outputs, compared with 103 for the Carrier
route and 127 for the Full pool; 69 parts have scores for both CADFit
and the Carrier route. CADFit failures include missing programs,
invalid STEP files and non-manifold predictions or references. On the
128 parts that have mesh inputs for both routes and an available
reference, CADFit scores .511, the Carrier route scores .617 and the
Full pool scores .934.

\paragraph{StepCAD.}
We run the public release in its 3-minute setting on an NVIDIA L4 GPU
with 12 vCPUs.

\paragraph{CADENA.}
CADENA~\cite{kabisov2026cadena} completes rollouts on all 477 available
inputs in about 3.48 GPU-hours on an RTX 4090, yielding 430 scorable
predictions. Its mean IND is .8050 among those predictions and .6495
across all 533 files. On the 400 single-solid files, the Full pool
reaches .8393 mean IND against .6510 for CADENA. CADENA achieves IND of
at least .999 on four parts, compared with 72 for the Carrier route
without merging and 188 for the Full pool.

\paragraph{Published comparisons.}
Willis et al.~\cite{willis2021fusion360} report 62.3\% and 67.5\% exact
recovery at 20 and 100 Fusion Gym steps on the 1,725 Gallery test
designs. Although we share that dataset, our kernel, action space and
budget unit differ. Zone graphs~\cite{xu2021zonegraph} extend each
target face to partition space into zones and run a learned best-first
search over extrusions that fill the inside zones. Their representation
covers 80\% of the 8,625 designs in the Gallery reconstruction dataset.
This figure measures coverage rather than a recovery rate within a
budget, and they evaluate their search on its own 440-design test set.
CADReasoner~\cite{kabisov2026cadreasoner} reports .491 mean IoU on the MCB benchmark
points after five greedy edits, while
CADEvolve~\cite{elistratov2026cadevolve} reports .514 and .552 for its
two RL models from rendered views; both protocols differ from IND.
RLCAD~\cite{yin2025rlcad} reads extrusions and revolves from B-rep
faces. Its latest version reports .900 IoU against .568 for the Gallery
agent on its own test set, while an earlier version reported .876
against .739 on 500 ABC B-reps. For that comparison, the RLCAD authors
extended the Gallery agent to support revolves and retrained it on
their own data.
BrepCoder~\cite{kim2026brepcoder} predicts DeepCAD command sequences
from B-reps, reaching 89.34\% command accuracy and 82.01\% parameter
accuracy.
Because we found no trained zone-graph, RLCAD or BrepCoder model that
we could run on our test sets, we cite their published numbers.
ParaCAD~\cite{qin2026paracad} generates parametric faces under its own
protocol.

\paragraph{Qualitative figure.}
Main-paper Fig.~3 shows single-solid ABC parts from the four
whole-file face-count tiers for a subset of the methods in main-paper
Table~1. We chose these parts from sheets of candidate parts; they are
examples, not a random sample. The value labeled IoU under each output
is IND. All renderers in a row use the source solid's camera, frame and
scale.

\section{Further related work}
\label{sec:supp_related_more}

\paragraph{Target-derived actions on B-reps.}
The Fusion~360 Gallery agent~\cite{willis2021fusion360} and zone graphs~\cite{xu2021zonegraph} derive candidate extrusions from the faces of the target.
RLCAD~\cite{yin2025rlcad} follows the same idea and adds revolves.
Zone graphs also extend every target face to partition space into zones.
CADParser~\cite{zhou2023cadparser} and Brep2Seq~\cite{zhang2024brep2seq} predict feature sequences from B-rep input.
We expand the tool set with holes, finishing operations (fillets, chamfers, and shells), and defeatured bodies.
Before any kernel replay, we screen Boolean combinations of these tools using inside/outside probe bits.

\paragraph{Self-training of shape programs.}
PLAD~\cite{jones2022plad} and SIRI~\cite{ganeshan2023siri} train a program inferrer on its own best programs.
Our expert iteration follows the same loop: it keeps only volume-exact programs and anchors the policy to its starting point with a KL term.
Alternating search and imitation also underlies DreamCoder~\cite{ellis2021dreamcoder} and expert iteration~\cite{anthony2017exit}.

\paragraph{Cell-complex reconstruction.}
Chauve et~al.~\cite{chauve2010robust} split space with detected planes and label each cell as inside or outside.
PolyFit~\cite{nan2017polyfit} intersects detected planes to form candidate faces and selects a watertight subset via binary linear programming.
Kinetic Shape Reconstruction~\cite{bauchet2020ksr} partitions space into convex cells and labels them with a min-cut, just as our mesh route does with curved carriers.
Both are limited to planes.
ComplexGen~\cite{guo2022complexgen} generates a full B-rep chain complex from points.
SPFN~\cite{li2019spfn} fits primitives to points with a supervised network.
Our mesh and point routes combine analytic fitting and set-prediction networks with a cell-complex min-cut.
Point2CAD~\cite{liu2024point2cad} and PlankAssembly~\cite{hu2023plankassembly} handle related inputs.
Point2Cyl~\cite{uy2022point2cyl} recovers extrusion cylinders from point clouds.

\paragraph{Three-view reconstruction and the visual hull.}
Classical methods reconstruct solids from orthographic views~\cite{wesley1981fleshing,sakurai1983solid,shin1998fast}.
The visual hull intersects silhouette cones~\cite{laurentini1994visual}.
Our drawing route belongs to this family: it intersects a box with the extruded view outlines, cuts the interior, and checks the result against the input lines by re-projection.

\paragraph{Metrics.}
CADENA~\cite{kabisov2026cadena} scores matches using a normal gate, and SPFN~\cite{li2019spfn} measures per-primitive coverage.
OSC builds on both.
NH-Rep~\cite{guo2022nhrep} shows that a Boolean tree over B-rep patches always exists, so volume exactness does not determine the boundary.
We therefore also report boundary and feature metrics (OSC and FeatR95).

\section{Results by part complexity}
\label{sec:supp_featstrat}
\begin{figure*}[t]
\centering
\includegraphics[width=\linewidth]{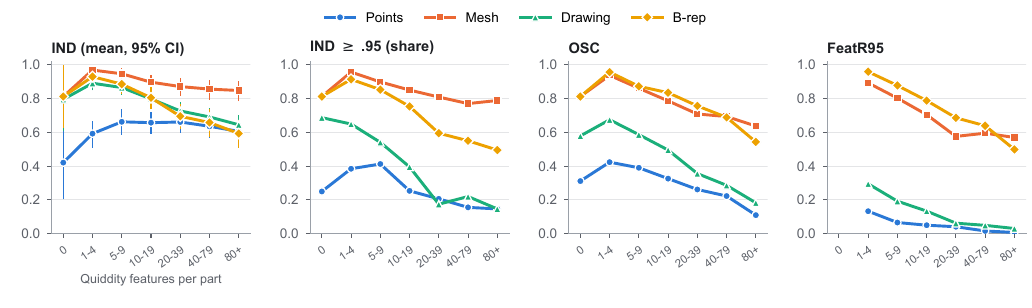}
\caption{\textbf{Our four routes by number of recognized features} on
the 748 parts of main-paper Table~1: mean IND with 95\% bootstrap
intervals, the share of parts with IND of at least .95, OSC and FeatR95.
Bins and counts as in Table~\ref{tab:supp_featstrat}.}
\label{fig:supp_featstrat_metrics}
\end{figure*}

\begin{figure*}[t]
\centering
\includegraphics[width=\linewidth]{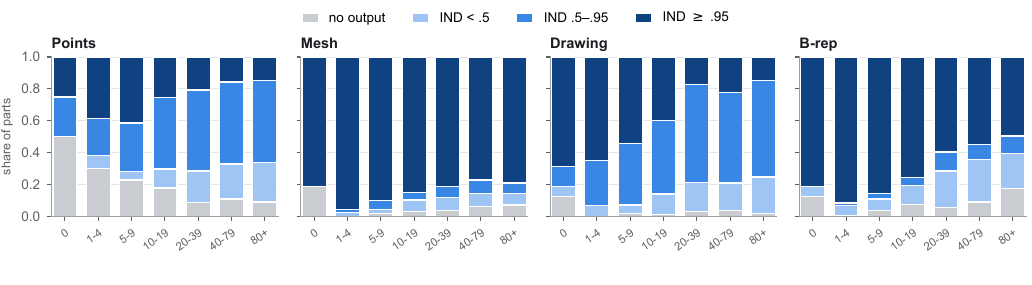}
\caption{\textbf{Outcomes by number of recognized features}: for each
route, the share of parts with no output, IND below .5, IND from .5 to
.95, and IND of at least .95.}
\label{fig:supp_featstrat_outcomes}
\end{figure*}

\begin{figure}[t]
\centering
\includegraphics[width=\columnwidth]{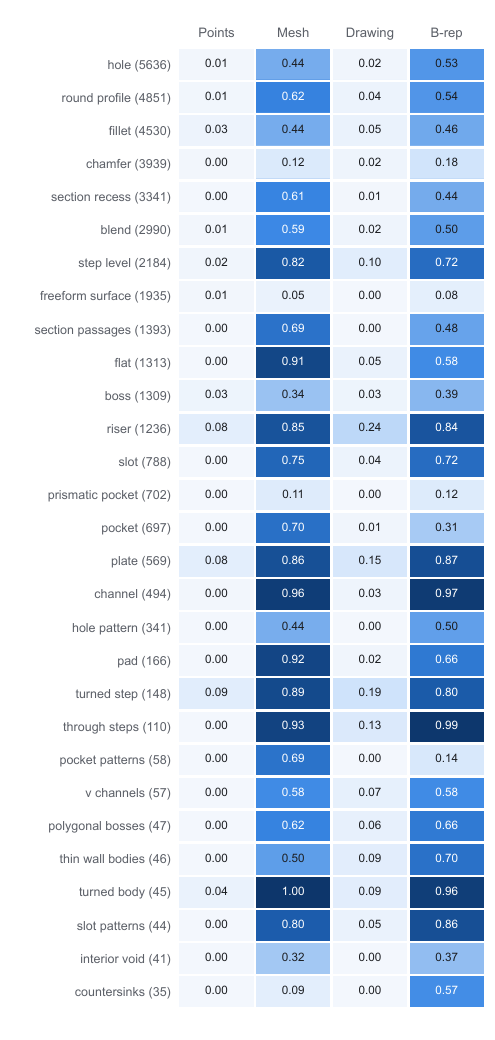}
\caption{\textbf{Recovered feature instances by family} on the 748
parts of main-paper Table~1. For each Quiddity family with at least 30
instances on the references (count in parentheses), the share of instances
that each route recovers: every supporting face is covered at least .95
at $\tau=10^{-3}$, as in FeatR95. Instances are pooled over parts; a
missing output recovers none.}
\label{fig:supp_featstrat_families}
\end{figure}

Figure~\ref{fig:supp_featstrat_metrics} and
Table~\ref{tab:supp_featstrat} group the 748 parts from main-paper
Table~1 by the number of features Quiddity recognizes on the reference,
using doubling bins.
Figure~\ref{fig:supp_featstrat_outcomes} splits each route's parts by
outcome: no output, IND below .5, IND from .5 to .95, and IND of at
least .95.
For the B-rep route, the share of parts with IND of at least .95 drops
from .91 at 1--4 features to .50 at 80 or more.
Up to 79 features, the B-rep route holds the highest or tied-highest
OSC and FeatR95 of the four routes; at 80 or more, the mesh route is higher (OSC .64 versus
.54, FeatR95 .57 versus .50).
The point and drawing routes stay below both in OSC and FeatR95 at
every complexity level.
Figure~\ref{fig:supp_featstrat_families} shows the share of feature
instances each route recovers by Quiddity family. Among families with at
least 1,000 instances, freeform surfaces and chamfers are the least
recovered for the B-rep and mesh routes.

\begin{table*}[t]\centering\footnotesize\setlength{\tabcolsep}{3.2pt}
\begin{tabular}{lr|cccc|cccc|cccc|cccc}
\toprule
 & & \multicolumn{4}{c}{Points} & \multicolumn{4}{c}{Mesh} & \multicolumn{4}{c}{Drawing} & \multicolumn{4}{c}{B-rep}\\
Features & $n$ & IND & $\geq$.95 & OSC & F95 & IND & $\geq$.95 & OSC & F95 & IND & $\geq$.95 & OSC & F95 & IND & $\geq$.95 & OSC & F95\\
\midrule
0 & 16 & .421 & .25 & .31 & -- & .812 & .81 & .81 & -- & .796 & .69 & .58 & -- & .812 & .81 & .81 & --\\
1--4 & 117 & .593 & .38 & .42 & .13 & .969 & .96 & .94 & .89 & .891 & .65 & .68 & .30 & .930 & .91 & .96 & .96\\
5--9 & 109 & .663 & .41 & .39 & .07 & .947 & .90 & .86 & .80 & .866 & .54 & .59 & .19 & .886 & .85 & .87 & .88\\
10--19 & 134 & .658 & .25 & .33 & .05 & .898 & .85 & .79 & .70 & .798 & .40 & .50 & .13 & .805 & .75 & .84 & .79\\
20--39 & 126 & .662 & .21 & .26 & .04 & .871 & .81 & .71 & .58 & .727 & .17 & .36 & .06 & .695 & .60 & .76 & .69\\
40--79 & 109 & .636 & .16 & .22 & .02 & .857 & .77 & .69 & .60 & .692 & .22 & .29 & .05 & .658 & .55 & .69 & .64\\
80+ & 109 & .606 & .15 & .11 & .01 & .848 & .79 & .64 & .57 & .645 & .15 & .18 & .03 & .594 & .50 & .54 & .50\\
\midrule
all valid & 720 & .632 & .26 & .29 & .05 & .897 & .85 & .77 & .69 & .772 & .36 & .44 & .13 & .764 & .70 & .78 & .74\\
\bottomrule
\end{tabular}
\caption{\textbf{Our four routes by number of recognized features.} We group the 748 single-solid ABC parts from main-paper Table~1 by the number of features Quiddity recognizes on the reference (the count that FeatR95 divides by). IND is the mean of the per-part IND values from main-paper Table~1, with failures scored as 0; because the ``all valid'' row averages over 720 parts rather than 748, its IND is higher than in main-paper Table~1. The $\geq$.95 column gives the share of parts with IND at least .95. OSC and FeatR95 include every part with a valid reference and score missing or failed outputs as 0, so every column covers the same parts. Main-paper Table~1 instead averages OSC and FeatR95 over each row's own eligible parts (Sec.~\ref{sec:supp_metrics}), giving slightly higher values than the ``all valid'' row. FeatR95 is undefined for parts with no recognized features. We exclude 28 parts whose reference has no valid face or could not be recognized.}
\label{tab:supp_featstrat}
\end{table*}

\section{Entity representation and program execution}
\label{sec:supp_exec}

All kernel operations, replays, and exports use Open CASCADE Technology
7.9.3 through the OCP 7.9.3.1 bindings, with CadQuery
2.8~\cite{occt,cadquery}. The extractor stores supporting surfaces,
oriented loops, edges, coedges, incidence, and per-instance identifiers.
Trimmed faces keep their analytic supports and boundaries. The extractor
can also record freeform supports, though the analytic builder and most
actions cannot construct them. The B-rep tools read face types, axes,
radii, areas, centroids, UV samples, and the convexity of shared edges.
For defeaturing and symmetry tests, they use whole-solid kernel
operations. Program construction resolves references against the
observed target. On a 512-part sample, extraction succeeded on every
B-rep. Rebuilding the stored representation preserved geometry and
boundary in all 512 cases, and 511 passed the topology check.

\section{Description-length diagnostics}
\label{sec:supp_mdl}
\begin{table}[t]
\centering
\small
\setlength{\tabcolsep}{3pt}
\begin{tabular}{@{}lrr@{}}
\toprule
Coding setup & Zero-to-CAD (771) & ABC (318) \\
\midrule
\multicolumn{3}{@{}l}{\emph{Two-part code of observed points}} \\
RANSAC + regularities & 5,547.6 & 2,617.0 \\
\quad With point slots & \textbf{2,343.3} & \textbf{1,210.1} \\
\quad With completion & 3,062.6 & 1,577.7 \\
\midrule
\multicolumn{3}{@{}l}{\emph{Conditional code of reference carriers}} \\
RANSAC + regularities & 395.9 & 167.2 \\
\quad With point slots & 378.5 & 159.3 \\
\quad With completion & \textbf{376.5} & \textbf{155.5} \\
\bottomrule
\end{tabular}
\caption{\textbf{Geometry coding at precision $\delta=.003$.}
Values are mean bits per part.
Each learned row adds one proposal source to the same analytic
baseline. Top: each code explicitly includes unexplained observations.
Bottom: each code conditions on the observations and encodes reference
carriers (the untrimmed surfaces underlying the faces).}
\label{tab:carrier_code}
\end{table}

\paragraph{Geometry.}
Our two-part code~\cite{rissanen1978mdl} charges
$L_\delta(E)+L_\delta(O\mid E)$ for carriers $E$ and observations $O$.
Greedy minimum description length selects carriers from a fixed pool at
precision $\delta$. The set code charges for types and parameters,
taking advantage of recurring directions, axes, and sizes. Each
unexplained point costs $3\log_2(1/\delta)$ bits in normalized
coordinates. Learned proposals mainly shorten the observation code by
leaving fewer points uncoded (Table~\ref{tab:carrier_code}).
Conditional carrier coding measures reference surfaces given the
observation and fitted predictions. Carriers paired with a prediction
use the predicted slot existence, type, and parameter uncertainty
estimated on separate folds, while unpaired carriers use an analytic
dictionary. At precision .003, the dictionary saves 283.7 bits per
Zero-to-CAD part and 102.1 bits per ABC part over the RANSAC-conditioned
code, and learned proposals shorten the code further.

\paragraph{Programs.}
Given a target $T$, a successful path $a_{1:k}$, and states $s_t$, the
conditional code length is $b_\pi=-\sum_t\log_2\pi(a_t\mid s_t,T)$
bits. Across 800 solved paths, six of seven policies show an
approximately linear relation between $\log_2 N$ (search expansions)
and $b_\pi$, with slopes .725--.795 and $R^2=.86$--$.93$, whereas depth
gives $R^2=.37$--$.55$. Offline distillation is the exception, with
slope .400. These fits describe known successful paths under the tested
policy and search. Code length depends on the action language and does
not estimate a global minimum. For our policy architecture, alternative
spellings and orderings of equivalent actions account for 2.63 bits
(95\% interval $[1.92,3.55]$) out of 6.91 bits per solved path.

\end{document}